\documentclass{article}

\usepackage[preprint]{neurips_2025}

\usepackage[utf8]{inputenc} 
\usepackage[T1]{fontenc}    
\usepackage{hyperref}       
\usepackage{url}            
\usepackage{booktabs}       
\usepackage{amsfonts}       
\usepackage{nicefrac}       
\usepackage{microtype}      
\usepackage{xcolor}         
\usepackage{amsmath}
\usepackage{enumitem}
\usepackage{wrapfig}
\usepackage{graphicx}
\usepackage{subcaption}
\usepackage{float}
\usepackage{multirow}

\usepackage{caption}

\usepackage{changepage} 
\usepackage{tabularx}

\usepackage{mathtools}

\usepackage{amssymb}
\usepackage{amsthm}
\usepackage{color}

\newcommand{\cut}[1]{{}}

\newcommand{\va}{{\mathbf{a}}}

\newcommand{\ve}{{\mathbf{e}}}

\newcommand{\vs}{{\mathbf{s}}}

\newcommand{\vv}{{\mathbf{v}}}

\newcommand{\vz}{{\mathbf{z}}}

\makeatletter
\let\@@span\span
\def\sp@n{\@@span\omit\advance\@multicnt\m@ne}
\makeatother

\newcommand{\bc}{\begin{center}}
\newcommand{\ec}{\end{center}}

\newcommand{\bdm}{\begin{displaymath}}
\newcommand{\edm}{\end{displaymath}}

\newcommand{\beq}{\begin{equation}}
\newcommand{\eeq}{\end{equation}}

\newcommand{\bfl}{\begin{flushleft}}
\newcommand{\efl}{\end{flushleft}}

\newcommand{\bt}{\begin{tabbing}}
\newcommand{\et}{\end{tabbing}}

\newcommand{\beqn}{\begin{align}}
\newcommand{\eeqn}{\end{align}}

\newcommand{\beqs}{\begin{align*}} 
\newcommand{\eeqs}{\end{align*}}  

\title{Towards General Computer Control with Hierarchical Agents and Multi-Level Action Spaces}

\author{%
  Zihan Dong\thanks{Equal contribution.} \\
  Workstation Solution \\
  Lenovo US \\
  Morrisville, NC 27560 \\
  \texttt{puma122707@gmail.com} \\
  \And
  Xinyu Fan\footnotemark[1] \\
  Information Science and Technology\\
  The University of Tokyo \\
  \And
  Zixiang Tang \\
  Advanced AI Technology Center \\
  Lenovo US \\
  Morrisville, NC 27560 \\
  \And
  Yunqing Li \\
  Advanced AI Technology Center \\
  Lenovo US \\
  Morrisville, NC 27560 \\
}

\begin{document}

\maketitle

\begin{abstract}
Controlling desktop applications via software remains a fundamental yet under‑served problem: existing multi‑modal large language models (MLLMs) ingest screenshots and task instructions to generate keystrokes and mouse events, but suffer from prohibitive inference latency, poor sample efficiency on long‑horizon sparse‑reward tasks, and infeasible on‑device deployment. We introduced a lightweight hierarchical reinforcement learning framework, \textbf{ComputerAgent}, that formulates OS control as a two‑level option process (manager/subpolicy), employs a triple‑modal state encoder (screenshot, task ID, numeric state) to handle visual and contextual diversity, integrates meta‑actions with an early‑stop mechanism to curb wasted interactions, and uses a compact vision backbone plus small policy networks for on‑device inference (0.015 B parameters). On a suite of 135 real‑world desktop tasks, ComputerAgent attains 92.1\% success on simple tasks (<8 steps) and 58.8\% on hard tasks ($\geq 8$ steps), matching or exceeding 200 B+ MLLM baselines on simple scenarios while reducing model size by over four orders of magnitude and halving inference time. Our results demonstrate that hierarchical RL offers a practical, scalable alternative to monolithic MLLM‑based automation for computer control.
\end{abstract}

\section{Introduction}\label{sec:intro}
\vspace{-4pt}

\paragraph{From siloed skills to software‐using agents.} Multi-modal Large Language Models (MLLM) such as GPT-4o, Claude 3, and Gemini 1.5 have matched—sometimes exceeded—human performance on language benchmarks, while foundation vision models rival experts on ImageNet\cite{schulman2017proximal} and COCO\cite{lin2014microsoft}.  Multimodal variants even process text, images, and speech in a single forward pass. Today, these `superhuman' models can accomplish a significant number of tasks in different domains, but challenges like \emph{see} a live desktop, \emph{understand} an instruction like `rename the latest invoice and email it', and \emph{act} through the graphical user interface (GUI) to complete the job remained. On the new OSWorld benchmark~\cite{xie2024osworldbenchmarkingmultimodalagents} (369 real tasks), humans succeed on $\approx72\%$, but the best published agent manages only $\approx12\%$.  Closing this gap would unlock \textbf{general‐purpose computer agents} able to automate workflows with transparent, click-for-click traces and \emph{no} private API dependence.

\paragraph{Why giant MLLMs may not be the answer.}  State‑of‑the‑art multimodal language models like GPT‑4o, Claude3, and Gemini1.5 boast impressive zero‑shot performance, but their sheer scale introduces several barriers to practical deployment.  First, pretraining such models demands vast, carefully curated multimodal corpora and weeks of multi‑GPU training, entailing millions of dollars in compute and human annotation costs. Second, inference requires GPUs with hundreds of gigabytes of memory to process high‑resolution screenshots in real time, ruling out on‑device usage in laptops or edge‑scale settings. Third, reliance on proprietary cloud APIs raises latency, data‑privacy, and regulatory hurdles in sensitive domains like healthcare and finance.  By contrast, a growing ecosystem of lightweight open‑source MLLMs (e.g., Llama3.2 Vision~\cite{lee2025efficient}) can be fine‑tuned and served on a single consumer GPU, albeit with some accuracy trade‑offs.

\begin{wrapfigure}{r}{0.54\textwidth}
  \centering
  \vspace{-1.0cm}
  \captionsetup{skip=3pt}    
  \includegraphics[width=0.4\textwidth]{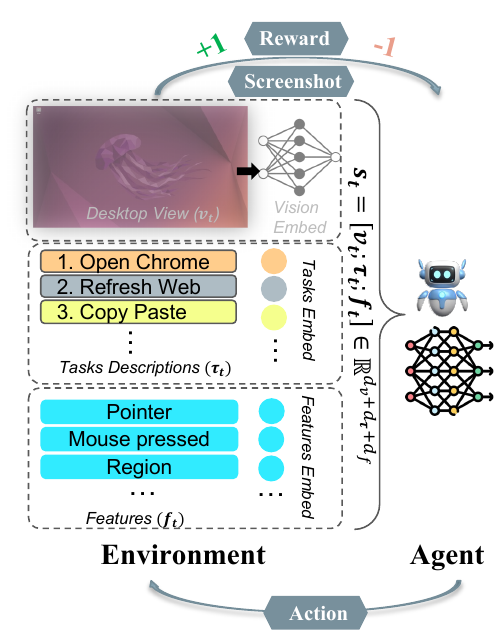}
  \caption{\textbf{Computer Agent}: Simulated System Screenshot, Task Description with ID as embedding and Feature States as input.}
  \vspace{-0.9cm}
  \label{fig:main}
\end{wrapfigure}

\paragraph{A lightweight approach: \emph{ComputerAgent}}
We therefore pursue a complementary direction: \emph{on-device} learning and inference using \textbf{ComputerAgent}, a hierarchical reinforcement-learning (HRL) framework that couples a compact vision encoder and two-level controller with a hybrid action space.  
ComputerAgent trains under sparse rewards, generalises across Windows, Ubuntu and macOS, and—critically—fits on a single consumer GPU, sidestepping both the monetary and privacy hurdles above.  
Its design choices are motivated by four challenges:  
(1) \emph{Sparse rewards} and \emph{long horizons} demand temporal abstraction;  
(2) \emph{Perceptual diversity} across OSs requires robust state embeddings;  (3) \emph{Text input} is indispensable for realistic tasks;  
(4) \emph{Enterprise deployment} calls for local inference.
We meet these challenges with:  

\begin{itemize}[leftmargin=*]
    \item \textbf{Hierarchical control.}  A manager–worker architecture decomposes the huge action space and mitigates sparse-reward credit assignment by converting 20-step problems into 3–5-step options.
    \item \textbf{Curriculum learning.}  Training progresses from 90 “easy’’ ($<\!8$-step) to 45 “hard’’ ($\ge\!8$-step) tasks, yielding policies that remain robust across the full difficulty range.
    \item \textbf{Early-stop meta-action.}  An explicit \texttt{EarlyStop} lets the agent terminate once the goal is satisfied, mirroring real-world usage and eliminating wasted actions.
    \item \textbf{Triple-modal state embedding.}  A compact fusion of screenshot, task goal and action context supplies the situational awareness needed for cross-OS generalisation.
\end{itemize}

\paragraph{Headline results.}
We defined 135 frequent OS tasks spanning copy–paste, web navigation, file management, and system settings. It executes with only $<1.15\times$ the human action count ($N_p/N_t$).  This work marks an initial step toward a locally deployable solution, offering a favorable trade‐off: ComputerAgent attains a better average accuracy of \textbf{80.8\%} versus \textbf{67.2\%} for pretrained MLLM‐based agents ~\cite{xie2024osworldbenchmarkingmultimodalagents}. Specifically, ComputerAgent reached to \textbf{92.1\% of human success on simple tasks} ($<8$ steps) and \textbf{72.2\% on hard tasks} ($\ge8$ steps. While the accuracy of ComputerAgent is slightly lower in accuracy, the demand for computational resources is significantly smaller (0.015 B vs.\ $>200$ B parameters MLLM), which makes the local deployment feasible and expandable.

\section{Related works}

\subsection{Existing computer agents}

Work on general computer agents splits into \emph{web} and \emph{desktop} tracks.
Web agents exploit HTML/DOM access: MiniWoB++ couples imitation with RL for 60 toy tasks \cite{humphreys2022learning}; Mind2Web expands to 2 000 real-site tasks but still relies on LLM reasoning over filtered pages \cite{deng2023mind2web}; WebGPT answers questions in a text-only browser, avoiding GUI control altogether \cite{nakano2021webgpt}.
Desktop agents pursue full OS control but remain heavyweight and cloud-oriented.  AssistGUI \cite{gao2024assistgui} and Microsoft’s UFO \cite{zhang2024ufo} drive Windows apps with vision encoders+GPT planners; Cradle \cite{tan2024cradle} scales to games and office suites using a large foundation model; OS-Copilot \cite{wu2024oscopilot} adds self-improvement yet still depends on scripted demos. OSWorld reveals fragility on open-ended tasks \cite{xie2024osworldbenchmarkingmultimodalagents}, while OmniParser explores lighter vision parsing without a control stack \cite{lu2024omniparserpurevisionbased}. Taken together, existing systems either leverage LLM templates with coarse actions or apply RL only in constrained web domains, leaving a gap for a trainable OS agent capable of executing more flexible, open-ended tasks with fine-grained control.


\subsection{Hierarchical reinforcement learning}
Temporal abstraction in RL originated with the \emph{options framework}, which formalizes reusable skills as semi-Markov actions with internal policies and termination rules~\cite{Sutton1999Options}. The Option-Critic architecture later showed how to learn both option policies and their stopping conditions end-to-end~\cite{Bacon2017OptionCritic}. Early deep variants such as h-DQN~\cite{Kulkarni2016HDQN} and Feudal Networks \cite{Vezhnevets2017Feudal} introduced explicit manager–worker hierarchies, enabling efficient exploration on sparse-reward Atari games~\cite{mnih2013playing}. Goal-conditioned methods like HIRO further improved data efficiency by off-policy training of both levels and automatic hindsight relabeling of sub-goals~\cite{Nachum2018HIRO}. Subsequent work explored stochastic latent skills~\cite{Florensa2017SNN} and timed sub-goals for non-stationary worlds (HiTS)~\cite{Gurtler2021HiTS}, while surveys provide comprehensive design guidelines~\cite{Pateria2021Survey,Narvekar2020CurriculumSurvey}.

Applying these ideas to pixel-level desktop control remains largely unexplored. A GUI exposes an unstructured visual state and a combinatorial hybrid action space, making it difficult for a high-level policy to discover meaningful sub-goals and for a low-level controller to receive formative rewards.  Existing attempts either rely on dense reward shaping or handcrafted sub-task hierarchies~\cite{Jain2022UIHRL}.  \textbf{ComputerAgent} bridges this gap with a domain-specific hierarchy: a lightweight high-level policy proposes discrete UI operations (e.g., \texttt{OpenMenu}, \texttt{Confirm}), while a low-level policy executes precise pointer trajectories and hot-key sequences.  Coupled with curriculum scheduling, this structure yields stable learning and fine-grained control without massive vision–language models.

\subsection{Self-termination and meta-control}

Temporal abstraction is only useful if an agent can decide \emph{when} to abandon a skill.
In the options framework~\cite{bacon2017option}, this is handled by a learned termination function~$\beta(s)$ that is jointly optimized with the option policy and the meta-controller \cite{Sutton1999Options,Bacon2017OptionCritic}.
Follow-up work treats termination as an independent critic that predicts long-term value if the option were to continue, improving stability and sample efficiency \cite{Harb2018TerminationCritic}.
Timed sub-goals enforce a fixed horizon for each option to combat non-stationarity (\textsc{HiTS}) \cite{Gurtler2021HiTS}, while ``learning to stop’’ techniques add an explicit \texttt{Stop} action to let the agent halt rollouts once it believes the task is solved \cite{Jiang2019LearningToStop}.
Yet most HRL systems still rely on exogenous signals such as episode timeout or environment failure; very few allow agents to \emph{wait}, \emph{pause}, or issue composite actions.

\textbf{ComputerAgent} extends the meta-action set with \texttt{Start}, \texttt{Stop}, \texttt{Wait}, and \texttt{Text\_Input}.
Because these decisions are trained end-to-end alongside the hierarchy, the agent can (i) proactively terminate an episode when the UI confirms success, (ii) insert waits to accommodate page loads, and (iii) bundle multi-character typing into a single high-level command—capabilities that prior computer control agents lack.
This learned meta-control closes the gap between option selection and full task management in open-ended desktop environments.

\subsection{Curriculum learning}
Sparse rewards and combinatorial action spaces often stall pixel-level RL. Curriculum learning and “reverse curriculum” mitigate this by exposing the agent to an (reversed) ordered sequence of tasks whose difficulty gradually increases \cite{Narvekar2020CurriculumSurvey,Florensa2017ReverseCurriculum}. Subsequent work automates syllabus design via progress-based sampling, yielding large gains in robotics and language games \cite{Matiisen2020TeacherStudent}.

We adopt a simple yet effective UI syllabus: begin with single-click targets on static screens, then introduce multi-step menu navigation, scrolling, text entry, and finally multi-application workflows. Task sampling is biased toward failure cases, ensuring the agent revisits skills it has not yet mastered. Combined with our hierarchical controller, this staged exposure accelerates convergence and boosts generalization to unseen desktop layouts.

\section{Hierarchical computer control}
\label{sec:Hierachical_Computer_Control}

In this work, we propose \textbf{ComputerAgent}, a hierarchical reinforcement learning framework designed for end-to-end computer control through GUI interactions. Our approach decomposes long-horizon tasks into high-level subgoal selection and low-level execution of fine-grained actions, with a rich state representation and a hybrid action space that closely mirrors human computer operation. In this section, we first introduce the hierarchical control architecture (Section~\ref{sec:Hierarchical_Control_in_Decision_Processes}), followed by the triple-modal state representation (Section~\ref{sec:Triple-Modal_State_Representation}) and the action space (Section~\ref{sec:hier_action}), and finally present the specialized reward and learning functions (Section~\ref{sec:Reward_Function_and_Learning_Objective}).

\begin{wrapfigure}{r}{0.4\textwidth}
  \centering
  \vspace{-0.9cm}
  \captionsetup{skip=3pt}    
  \includegraphics[width=0.4\textwidth]{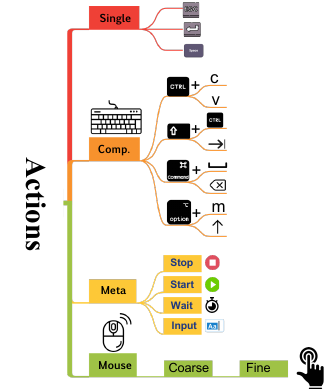}
  \caption{Action Hierarchy.}
  \vspace{-0.4cm}
  \label{fig:hier_action}
\end{wrapfigure}

\subsection{Hierarchical control in decision processes}
\label{sec:Hierarchical_Control_in_Decision_Processes}

To handle the complex and temporally extended tasks in computer control, we design our agent with two distinct levels of control 
as shown in Figure~\ref{fig:hier_action}
: the high-level controller selects macro-actions (or subgoals) that reflect semantic task progress, while the low-level controller executes precise actions (e.g., mouse clicks, key presses) to realize these subgoals. We denote the overall policy as a composition:
\[
\pi(\va_t \mid \vs_t) = \pi_H(\va_t^H \mid \vs_t) \cdot \pi_L(\va_t^L \mid \vs_t, \va_t^H),
\]
where \(\vs_t \) is the state at time \( t \), \(\va_t^H \) is the macro-action chosen by the high-level policy, and \(\va_t^L \) is the corresponding low-level action sequence. 


Specifically, our action space follows a hierarchical decision-making framework. As shown in Figure~\ref{fig:hier_action}, the high-level action consists of four macro-actions: single key, composite key, meta-action, and mouse control. The low-level action, in contrast, comprises more fine-grained micro-actions. Details of the action hierarchy are provided in Section~\ref{sec:hier_action} and derivations and hyperparameter settings in Appendix~\ref{appendix:vb_pb_learning_obj}.

\subsection{Triple-modal state representation}
\label{sec:Triple-Modal_State_Representation}
To ground decision making in the current image context and progress history, we propose a \emph{triple embedding scheme} that fuses three complementary modalities:

\begin{itemize}[leftmargin=1.0em]
    \item \textbf{Visual embedding:} A vision backbone holder\( f_v \), designed to match the output dimensionality of WideResNet50~\cite{he2016deep}, is used in this Phase-1 implementation to process the screenshot holders at time \( t \):
    \[
    \vv_t = f_v(\texttt{image}_t) \in \mathbb{R}^{1024}.
    \]
    This preserves compatibility with future upgrades to pretrained visual encoders such as WideResNet50, yielding a 1024-dimensional output.

    \item \textbf{Task embedding:} A task encoder \( f_t \) maps the task description into a semantic vector. We experiment with discrete task ID and dense embeddings from OpenAI's Ada-002 model (1536 dimensions)~\cite{OpenAI_Introducing_Operator}. To reduce dimensionality, we apply PCA-based compression~\cite{jolliffe2002principal}:
    \[
    \ve_t = f_t(\texttt{task}) \in \mathbb{R}^{d_t}, \qquad \text{where} \quad \ve_t = W_{\text{PCA}}^\top (\vz_t - \bar{\vz}),
    \]
    where \( \vz_t \in \mathbb{R}^{1536} \) is the original Ada-002 embedding, \( \bar{\vz} \) is the mean vector, and \( W_{\text{PCA}} \in \mathbb{R}^{1536 \times d_t} \) contains the top \( d_t \in \{64, 128\} \) principal components.

    \item \textbf{State embedding:} A simple feed-forward encoder \( f_s \) processes low-dimensional structured state information. Let \( P \) denote the pointer position, \( n \) the current step count, \( R \) the current region, \( S \) the current subregion, and \( I_m \) the mouse press state. The combined state is encoded as:
    \[
    \vs'_t = f_s([P, n, R, S, I_m]) \in \mathbb{R}^{d_s},
    \]
    where \( d_s \) is the learned dimensionality of the state embedding. This compact representation captures the agent’s internal execution state at each step.

\end{itemize}

These embeddings are concatenated to form the unified state representation:
\[
\vs_t = \big[ \vv_t; \ve_t; \vs'_t \big] \in \mathbb{R}^{d_v + d_t + d_s'}. 
\]
This representation is used as input for both the Q-networks (in DQN~\cite{mnih2013playing}) and policy/value networks (in PPO~\cite{schulman2017proximal} and A2C~\cite{mnih2016asynchronous}).

\subsection{Hierarchical action space}
\label{sec:hier_action}
Our action space employs a hierarchical structure as shown in Figure~\ref{fig:hier_action}, where we split the whole action space into four macro-actions followed by a series of corresponding fine-grained micro-actions. The agent’s action space is structured hierarchically as \(\mathcal{A} = \mathcal{A}_{\text{type}} \times \mathcal{A}_{\text{content}}\), where \(\mathcal{A}_{\text{type}} = \left\{ \mathcal{A}_{\text{single}} \cup \mathcal{A}_{\text{hot}} \cup \mathcal{A}_{\text{meta}} \cup \mathcal{A}_{\text{mouse}} \right\}\) denotes the set of action modes, and \(\mathcal{A}_{\text{content}}\) is the set of concrete actions available within each mode. The subspaces have cardinalities \(|\mathcal{A}_{\text{single}}| = 92\), \(|\mathcal{A}_{\text{hot}}| = 78\), and \(|\mathcal{A}_{\text{meta}}| = 4\). The mouse control actions are defined compositionally as \(\mathcal{A}_{\text{mouse}} = \mathcal{R} \times \mathcal{S} \times \mathcal{I}_m\), where \(|\mathcal{R}| = 9\), \(|\mathcal{S}| = 9\), and \(|\mathcal{I}_m| = 8\), with \(\mathcal{R}\) representing coarse regions, \(\mathcal{S}\) fine-grained subregions, and \(\mathcal{I}_m\) interaction types (e.g., \texttt{left\_press}, \texttt{left\_release}) See App.~\ref{Agent_Function_Examples}. At each timestep, the agent selects a mode from \(\mathcal{A}_{\text{type}}\) and executes an action within the corresponding \(\mathcal{A}_{\text{content}}\) space. The hierarchical structure of our action space enables the agent to flexibly and efficiently mimic human behavior during computer interaction. The detailed descriptions are further explained in App.~\ref{appendix:hier_action}.

\subsection{Reward function and learning objective}
\label{sec:Reward_Function_and_Learning_Objective}
\textbf{Reward function.} At timestep \(t\) the total reward combines manager–level guidance, fine‑grained subpolicy feedback, and explicit penalties:
\begin{equation}
  r_t \;=\;
      R_{\mathrm{mgr}}\!\bigl(\vs_t,\va_t^{\mathrm{type}}\bigr)
      \;+\;
      \alpha\,
      R_{\mathrm{sub}}\!\bigl(\vs_t,\va_t^{\mathrm{content}}\bigr)
      \;-\;
      P\!\bigl(\vs_t,\va_t\bigr),
  \label{eq:total_reward}
\end{equation}
where \(\va_t =
      \bigl(\va_t^{\mathrm{type}},\va_t^{\mathrm{content}}\bigr)
      \in \mathcal{A}_{\mathrm{type}}\times\mathcal{A}_{\mathrm{content}}\).
\(P(\cdot)\) aggregates penalties for repeated actions, idle steps, early termination, \emph{etc},
and \(\alpha>0\) balances hierarchical levels.

\textbf{Manager reward.} The manager is rewarded only when its selected macro‑action matches the ground‑truth sequence:
\begin{equation}
  R_{\mathrm{mgr}}
  \;=\;
  \bigl(r_{\mathrm{corr}} + \beta\,r_{\mathrm{streak}}\bigr)\,
  \mathbb{1}\!\bigl[\va_t^{\mathrm{type}} = \va_t^{\star}\bigr],
  \label{eq:manager_reward}
\end{equation}
where \(\va_t^{\star}\) is the target macro‑action, \(r_{\mathrm{corr}}>0\) is the base bonus,
\(r_{\mathrm{streak}}\) grows with consecutive correct selections, and \(\beta\) scales the streak bonus.

\paragraph{Subpolicy reward.}
The intrinsic reward depends on the specific micro‑action modality.

\begin{enumerate}[leftmargin=1.2em,itemsep=4pt]
  \item \emph{Mouse actions.}\;%
        Let \((r,s,i_m)\in\mathcal{R}\times\mathcal{S}\times\mathcal{I}_m\) denote the chosen region, sub‑region, and interaction, and
        \((r^\star,s^\star,i_m^\star)\) the ground truth.
        \begin{equation}
          R_{\mathrm{sub}}
          \;=\;
          \underbrace{\Bigl(1-\tfrac{d}{d_{\mathrm{th}}}\Bigr)_{+}\,r_{\mathrm{reg}}}_{R_{\mathrm{pos}}}
          \;+\;
          \underbrace{r_{\mathrm{int}}\,
          \mathbb{1}[i_m=i_m^\star]}_{R_{\mathrm{act}}},
          \qquad
          d=\lVert(r,s)-(r^\star,s^\star)\rVert_{2}.
          \label{eq:mouse_reward}
        \end{equation}
        Here \((x)_{+}\!=\max(0,x)\) enforces the positional bonus to decay linearly to zero beyond distance \(d_{\mathrm{th}}\);
        \(r_{\mathrm{reg}}, r_{\mathrm{int}}>0\) weight spatial accuracy and interaction correctness, respectively.

  \item \emph{Keyboard and hot‑key actions.}\;%
        For any discrete key sequence \(k\in\mathcal{A}_{\mathrm{content}}\),
        \begin{equation}
          R_{\mathrm{sub}}
          \;=\;
          r_{\mathrm{key}}\,
          \mathbb{1}[k = k^\star],
          \label{eq:key_reward}
        \end{equation}
        with \(k^\star\) the expected key and \(r_{\mathrm{key}}>0\) the unit reward.
\end{enumerate}

Equations~\eqref{eq:total_reward}–\eqref{eq:key_reward}
align with the hierarchical action definition
\(\mathcal{A}=\mathcal{A}_{\mathrm{type}}\times\mathcal{A}_{\mathrm{content}}\)
and present indicator‑based rewards compactly without redundant keystroke examples.

\textbf{Penalty function.} To discourage repeated errors, idleness, inefficient exploration, and premature termination, the total penalty is the additive composition
\begin{align}
  P(\vs_t,\va_t)
  \;=&\;
    \underbrace{\lambda_{\mathrm{rep}}\,
      \mathbb{1}\!\bigl[\va_t=\va_{t-1}\bigr]}_{P_{\mathrm{repeat}}}
    \;+\;
    \underbrace{\lambda_{\mathrm{idle}}\,
      \mathbb{1}\!\bigl[\va_t^{\mathrm{type}}=\textsc{Wait}\bigr]}_{P_{\mathrm{pointer}}}
    \;+\;
    \underbrace{\lambda_{\mathrm{step}}}_{P_{\mathrm{step}}}
    \notag\\[2pt]
    &\;+\;
    \underbrace{\lambda_{\mathrm{early}}\,
      \mathbb{1}\!\bigl[t<T^{\star}\wedge\va_t^{\mathrm{type}}=\textsc{Stop}\bigr]}_{P_{\mathrm{early\_end}}}
    \;+\;
    \underbrace{\lambda_{\mathrm{exp}}\,
      \bigl(d_{\mathrm{explore}}-d_{\min}(\vs_t)\bigr)_{}}_{P_{\mathrm{explore}}},
  \label{eq:penalty}
\end{align}
where all coefficients \(\lambda_{\bullet}>0\) and  
\((x)_{+}=\max(0,x)\).  
Here \(d_{\min}(\vs_t)\) is the distance from the current mouse position to the nearest task‑relevant target, and \(T^{\star}\) is the minimum feasible episode length.  
Appendix~\ref{app:Penalty} details the coefficient choices and function definations.

\section{Experiment}
\label{sec:exp}
In this section, we talked about the data collection process (Section~\ref{sec:dataset_and_pipleline}), evaluated the performance of ComputerAgent using three reinforcement learning methods (PPO, A2C, and DQN), highlighting the superior effectiveness of DQN (Section~\ref{sec:main_result}). Additionally, we conduct several ablation studies and downstream analyses to uncover the working mechanisms of our framework (Section~\ref{sec:ablation_study}).

\subsection{Experimental settings}

\label{sec:dataset_and_pipleline}
\textbf{Dataset and Pipeline:} To build our task suite, we designed 135 practical tasks in total, of which 90 as “simple”  \(N_s\) and 45 as “hard” \(N_h\). To support this curriculum, we categorize tasks into two difficulty levels: \textit{easy} (action sequence length \(< 8\)) and \textit{hard} (length \(\ge 8\)). More detailed steps of data collection can be found in Appendix~\ref{app:Data}. We employ a smooth-decay curriculum strategy by defining the task sampling probabilities as follows: 
\[p_{\text{simple}}(t) = 1 - \left(\frac{t}{T}\right)^\alpha; p_{\text{hard}}(t) = 1 - p_{\text{simple}}(t)\]
where \(t \in [0, T]\) is the normalized training progress and \(\alpha > 0\) controls the decay rate. To ensure that each group receives training episodes proportional to its size, we define \(N_s = |\{t \in \text{TASKS}: |\text{action\_seq}(t)| < 8\}|\) and \(N_h = |\{t \in \text{TASKS}: |\text{action\_seq}(t)| \ge 8\}|\), and enforce:
\[
k\frac{\#\text{episodes}_s}{\#\text{episodes}_h} = \frac{N_s}{N_h}.
\]

In our design, the expected proportion of hard-task episodes over the training horizon is
\[
p_h = \int_{0}^{1} t^{\alpha}\,\mathrm{d}t = \frac{1}{\alpha + 1}.
\]
To match group proportions, we set
\[
\frac{1}{\alpha + 1} = \frac{N_h}{N_s + N_h} \quad \Longrightarrow \quad \alpha = \frac{N_s}{N_h}.
\]
Thus, by setting \(\alpha = \frac{N_s}{N_h}\), the sampling smoothly transitions from favoring easy tasks early (\(t \approx 0\)) to hard tasks later (\(t \approx 1\)), while ensuring balanced exposure across difficulty levels.

\label{Experiment_settings}

\textbf{Model Settings:} For all experiments, we adopt a triple‑modal state representation comprising a vision backbone with a 1‑layer MLP into $\mathbb{R}^{1024}$, a \textit{Task} embedding formed by concatenating a learnable task‐ID embedding and one MLP‑encoded description (32+32=64 dims), and a 5‑dimensional numeric state encoded via a 2‑layer MLP into $\mathbb{R}^{64}$. These features are concatenated and fed through a 4‑layer fully connected policy/value network (512 hidden units) to produce Q‑values or action logits. Agents train for 100\,000 episodes with sub-policy rewards scaled by $\alpha=1.2$ and streak bonuses of 5 for the manager and 1 for sub-policies. Because our practical tasks rarely exceed 100 steps, we cap each episode at 100 steps. However, thanks to the model’s built-in self-termination mechanism, ComputerAgent can end an episode early whenever it has completed the task. Hyperparameters and implementation details in Appendix~\ref{app:Exp_setup}.

\textbf{Evaluation and Infrastructure Settings:} We report Precision, Recall, and F1 = 2·P·R/(P+R) over atomic GUI actions (see Appendix~\ref{app:metrics} for TP/FP/FN definitions). The infrastructure requirements for replication are detailed in Appendix~\ref{app:infrastructure}.

\subsection{Main results}
\label{sec:main_result}
In this section, each experiment was carried out in three independent runs, and the results are reported as mean and standard deviation to ensure statistical reliability and reflect overall models performance.

\textbf{Embedding Effectiveness.} As shown in Table~\ref{tab:combined_results}, our triple-modal embedding produces substantial gains in all agents. For DQN, precision is 70.4\% vs 80.8\% (+14.8\%), and normalized reward is 82.6\% vs 89.5\% (+8.4\%). For PPO, recall is 56.2\% vs 67.7\% (+36.5\%), F1 is 56.0\% vs 65.6\% (+17.1\%), and reward is 56.1\% vs 65.6\% (+16.9\%). For A2C, precision is 72.4\% vs 78.0\% (+7.6\%), recall is 45.2\% vs 53.6\% (+18.6\%), F1 is 48.2\% vs 56.2\% (+16.6\%), and reward is 40.1\% vs 49.0\% (+22.2\%). These consistent improvements confirm that embedding rich task context substantially enhances decision accuracy, coverage, and sparse-reward learning. For further analysis of the impact of task difficulty, see Appendix~\ref{app:main_result}.

\begin{figure*}[htbp]
  \centering
  \begin{subfigure}[b]{0.54\linewidth}
    \includegraphics[width=\linewidth]{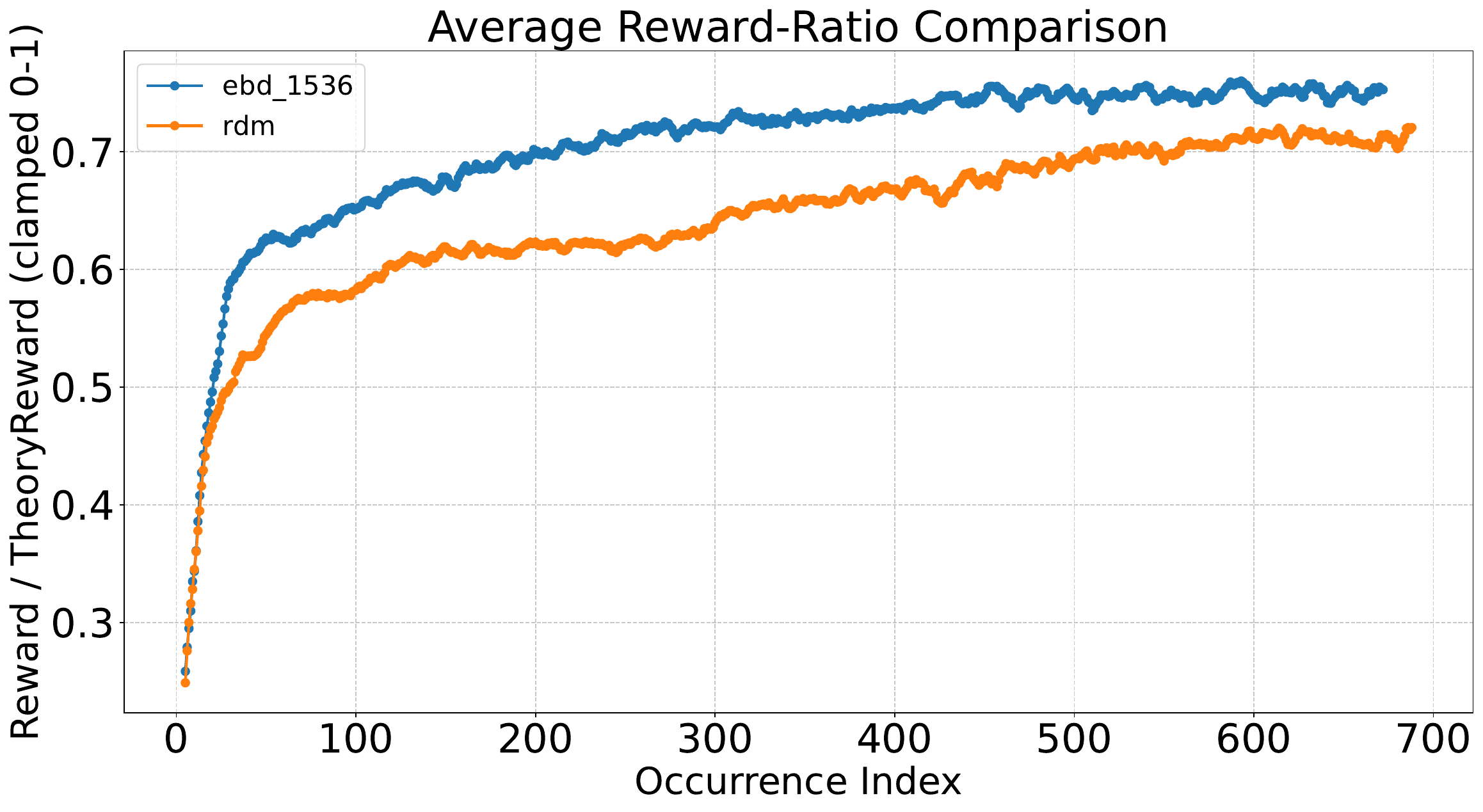}
    \caption{Curriculum Learning}
    \label{fig:curriculum}
  \end{subfigure}\hspace{0pt}%
  \begin{subfigure}[b]{0.46\linewidth}
    \includegraphics[width=\linewidth]{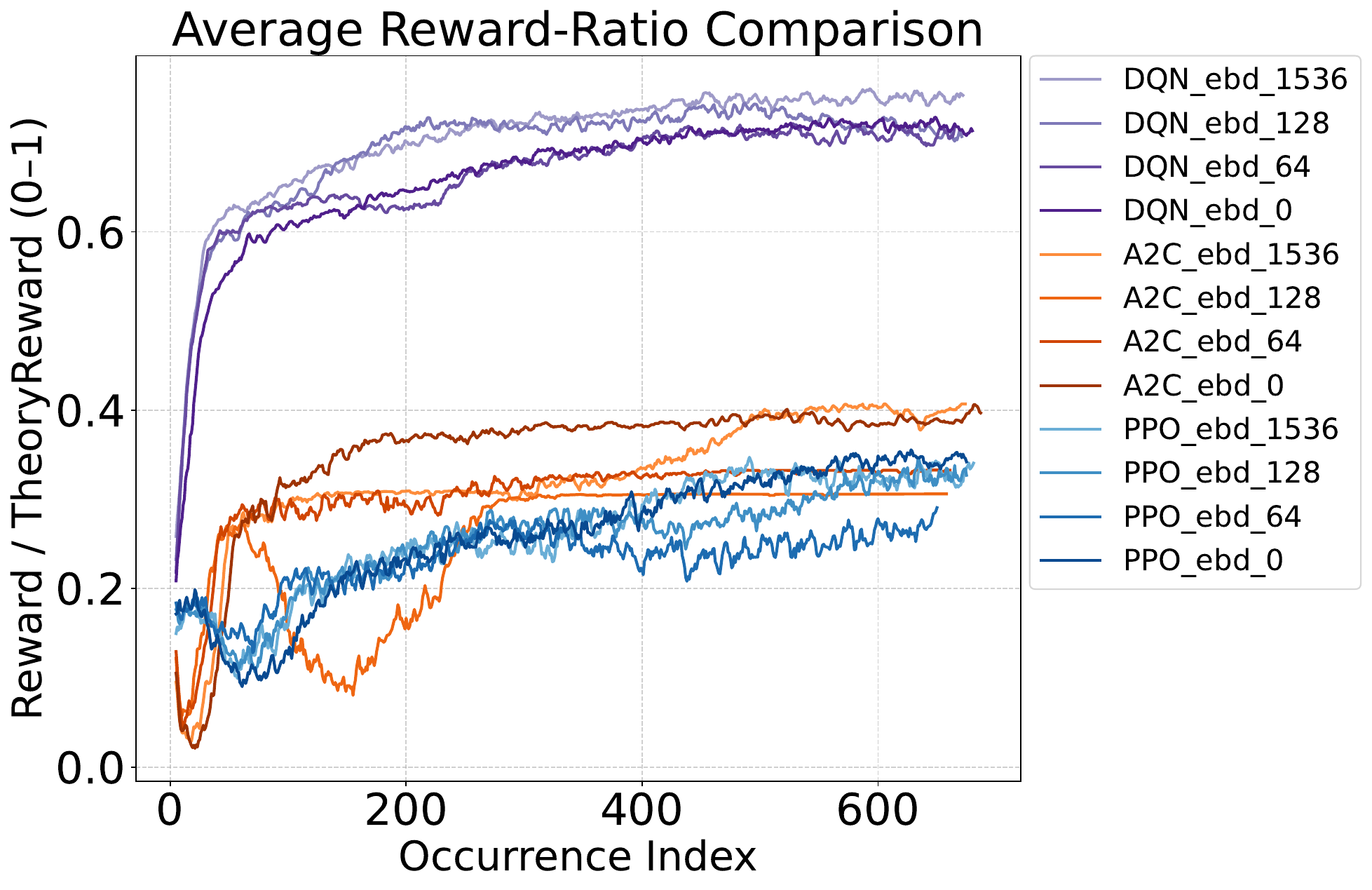}
    \caption{Embedding Dimension}
    \label{fig:embedding_dim}
  \end{subfigure}\hspace{0pt}%
  \caption{(a) Curriculum learning with a 1536‑dimensional embedding ($ebd_{1536}$, the blue curve) and random task sampling (b)Average Reward‑Ratio Comparison of PPO, A2C, and DQN agents using embedding dimensions of 0, 64, 128, and 1536 over 700 task occurrences.}
  \label{fig:Exp_Result}
  \vspace{-0.3cm}
\end{figure*}

\begin{table}[H]
  \centering
  \small
  \setlength{\tabcolsep}{0.5pt}      
  \renewcommand{\arraystretch}{1.1} 
  \begin{adjustwidth}{-0.05cm}{-0.05cm}
    \vspace{-0.3cm}
    \caption{Presents results across 5 experiment categories: embedding dimension, hierarchical comparison, embedding type, curriculum learning, and mouse control. “dim” indicates the embedding dimension evaluated; “Base” denotes the baseline without hierarchical structure; “CA” refers to our model, computer agent; “raw” combines raw task ID, dummy image, and meta information; “no\_id” omits the task ID embedding; “rdm” represents random curriculum sampling; and “ms” corresponds to the multi‑step mouse control variant. OSWorld with GPT-4o is considered the Baseline for comparison. Normalized reward is scaled from 0 to 1 to facilitate comparison across settings.}
    \vspace{+0.2cm}
  \begin{minipage}[t]{0.5\textwidth}
    \begin{tabular}{lcccc}
      \toprule
      \multicolumn{5}{l}{\textbf{Embedding Type (no\_id, id\_w/o\_ebd vs.\ CA)}} \\
      Model (dim) & Prec  & Rec  & F1   & N. Rwd \\
      \midrule
        $DQN_{raw}$     & $70.4_{\pm1.22}$   & $75.6_{\pm1.45}$   & $72.5_{\pm1.30}$   & $82.6_{\pm0.33}$  \\
        $DQN_{no-id}$   & $63.1_{\pm0.86}$  & $72.3_{\pm2.81}$   & $66.3_{\pm0.89}$  & $70.5_{\pm2.74}$   \\
        $PPO_{raw}$     & $75.3_{\pm0.67}$  & $56.2_{\pm1.66}$   & $56.0_{\pm1.85}$   & $56.1_{\pm7.71}$   \\
        $PPO_{no-id}$   & $66.0_{\pm1.20}$   & $66.5_{\pm0.13}$  & $63.6_{\pm0.78}$  & $64.1_{\pm1.03}$   \\
        $A2C_{raw}$     & $72.4_{\pm1.01}$   & $45.2_{\pm0.88}$  & $48.2_{\pm0.85}$  & $40.1_{\pm1.11}$   \\
        $A2C_{no-id}$   & $74.5_{\pm0.06}$ & $46.9_{\pm1.00}$   & $49.6_{\pm0.12}$  & $40.0_{\pm0.67}$  \\
        
      \midrule
      \multicolumn{5}{l}{\textbf{Hierarchical (Flat structure vs.\ CA)}} \\
      DQN         & $71.2_{\pm1.17}$ & $64.4_{\pm0.06}$ & $65.3_{\pm0.71}$ & $59.5_{\pm0.64}$ \\
      PPO         & $70.5_{\pm1.57}$ & $60.5_{\pm0.27}$ & $59.8_{\pm0.84}$ & $54.4_{\pm0.72}$ \\
      A2C         & $64.2_{\pm2.31}$ & $48.3_{\pm0.44}$ & $49.4_{\pm1.10}$ & $44.2_{\pm0.17}$ \\

      \midrule
      \multicolumn{5}{l}{\textbf{Curriculum (random vs.\ CA)}} \\
      $DQN_{rdm}$ & $78.3_{\pm2.26}$ & $79.6_{\pm2.21}$ & $78.9_{\pm2.27}$ & $88.0_{\pm0.04}$ \\
      $PPO_{rdm}$ & $65.0_{\pm1.30}$ & $64.5_{\pm8.09}$ & $61.0_{\pm5.71}$ & $60.4_{\pm10.70}$ \\
      $A2C_{rdm}$ & $70.2_{\pm0.65}$ & $48.4_{\pm2.62}$ & $50.2_{\pm2.96}$ & $41.6_{\pm3.68}$ \\

      \midrule
      \multicolumn{5}{l}{\textbf{Sub-region Control (w/o subr vs.\ CA)}} \\ 
        $DQN_{ms}$     & $78.7_{\pm0.52}$  & $80.1_{\pm0.38}$  & $79.2_{\pm0.44}$  & $87.9_{\pm0.03}$ \\
        $PPO_{ms}$     & $68.6_{\pm1.54}$  & $64.5_{\pm1.66}$  & $63.3_{\pm0.88}$  & $61.6_{\pm1.57}$ \\
        $A2C_{ms}$     & $74.8_{\pm0.89}$  & $48.0_{\pm0.84}$  & $50.1_{\pm0.99}$  & $41.3_{\pm0.28}$ \\
      \bottomrule
    \end{tabular}
  \end{minipage}\hfill
  \begin{minipage}[t]{0.5\textwidth}
    \begin{tabular}{lcccc}

      \toprule
      \multicolumn{5}{l}{\textbf{Embedding Dimension Study} 
  (Cont.)} \\
      Model (dim) & Prec  & Rec  & F1   & N. Rwd \\
      \midrule
        $DQN_{0}$      & $79.2_{\pm1.30}$  & $79.9_{\pm1.50}$  & $79.5_{\pm1.40}$  & $88.0_{\pm0.70}$ \\
        $PPO_{0}$      & $71.0_{\pm1.09}$  & $63.8_{\pm1.75}$  & $63.8_{\pm1.19}$  & $62.8_{\pm2.88}$ \\
        $A2C_{0}$      & $78.2_{\pm5.68}$  & $53.4_{\pm3.05}$  & $56.1_{\pm4.74}$  & $48.2_{\pm3.36}$ \\[0.5ex]
        $DQN_{64}$     & $78.0_{\pm1.34}$  & $79.5_{\pm0.43}$  & $78.6_{\pm1.03}$  & $87.8_{\pm0.06}$ \\
        $PPO_{64}$     & $67.1_{\pm0.19}$  & $66.4_{\pm2.30}$  & $64.3_{\pm1.76}$  & $64.6_{\pm4.75}$ \\
        $A2C_{64}$     & $74.6_{\pm0.98}$  & $49.3_{\pm3.81}$  & $51.9_{\pm3.06}$  & $43.9_{\pm5.28}$ \\[0.5ex]
        $DQN_{128}$    & $77.9_{\pm2.11}$  & $79.6_{\pm1.10}$  & $78.6_{\pm1.72}$  & $87.3_{\pm1.74}$ \\
        $PPO_{128}$    & $71.2_{\pm5.96}$  & $63.3_{\pm4.17}$  & $62.2_{\pm3.00}$  & $65.0_{\pm0.52}$ \\
        $A2C_{128}$    & $74.9_{\pm3.71}$  & $49.5_{\pm5.25}$  & $52.6_{\pm4.15}$  & $44.3_{\pm7.13}$ \\[0.5ex]
        $DQN_{1536}$   & $80.8_{\pm0.98}$  & $82.0_{\pm0.32}$  & $81.3_{\pm0.70}$  & $89.5_{\pm1.81}$ \\
        $PPO_{1536}$   & $67.8_{\pm0.43}$  & $67.7_{\pm2.86}$  & $65.6_{\pm1.47}$  & $65.6_{\pm5.63}$ \\
        $A2C_{1536}$   & $78.0_{\pm5.11}$  & $53.6_{\pm1.45}$  & $56.2_{\pm2.94}$  & $49.0_{\pm1.12}$ \\
        
      \midrule
      \multicolumn{5}{l}{\textbf{Computer Agent (Our model)}} \\
        $DQN_{CA}$    & \textcolor{brown}{$80.8_{\pm0.98}$}  & \textcolor{brown}{$82.0_{\pm0.32}$}  & \textcolor{brown}{$81.3_{\pm0.70}$}  & \textcolor{brown}{$89.5_{\pm1.81}$} \\


        $PPO_{CA}$    & $67.8_{\pm0.43}$  & $67.7_{\pm2.86}$  & $65.6_{\pm1.47}$  & $65.6_{\pm5.63}$ \\
        $A2C_{CA}$    & $78.0_{\pm5.11}$  & $53.6_{\pm1.45}$  & $56.2_{\pm2.94}$  & $49.0_{\pm1.12}$ \\
      \midrule
      \multicolumn{5}{l}{\textbf{OSWorld}} \\
      GPT-4o     & $65.6_{\pm0.14}$ & $57.9_{\pm4.18}$ & $58.0_{\pm3.38}$ & $35.2_{\pm0.43}$ \\
      GPT-4.1     & $67.2_{\pm0.84}$ & $61.0_{\pm0.14}$ & $60.8_{\pm0.49}$ & $39.4_{\pm1.7}$ \\
      \bottomrule
    \end{tabular}
  \end{minipage}

  \label{tab:combined_results}
  \end{adjustwidth}
\end{table}
\vspace{-0.4cm}

\textbf{Hierarchical Control.}  
Replacing a flat action space with our manager–worker hierarchy yields substantial normalized‑reward gains: DQN increases from 59.5\% to 89.5\% (+50.4\%), PPO from 54.4\% to 65.6\% (+20.6\%), and A2C from 44.2\% to 49.0\% (+10.9\%). This vindicates temporal abstraction for mitigating sparse‑reward credit assignment in long‑horizon image tasks.

\textbf{Curriculum Learning.}  
Under DQN, our fixed “easy‐to‐hard” curriculum achieves a normalized reward of 89.5\%, compared to 88.0\% under random task sampling. Similar, though smaller, gains are observed for PPO (65.6\% vs.\ 60.4\%) and A2C (49.0\% vs.\ 41.6\%). Moreover, curriculum learning converges substantially faster and with lower variance, accelerating early reward accumulation and yielding smoother progression across the full difficulty spectrum (Figure~\ref{fig:curriculum}). While random sampling eventually reaches comparable plateau levels, it exhibits greater instability on mid‐difficulty tasks—an issue our curriculum effectively mitigates.

\textbf{Embedding Dimension.}
Figure~\ref{fig:embedding_dim} presents reward‐vs‑episode curves for embedding dimensions $\{0,64,128,1536\}$.  DQN with a 1536‑dimensional embedding attains the highest normalized reward ratio ($\approx$ 89.5\%), outperforming both its lower‑dimensional variants and all PPO/A2C configurations. Across all algorithms, increasing embedding size accelerates convergence and raises peak performance, underscoring the value of rich state representations in sparse‑reward RL tasks.

DQN peaks at 1536 dimensions (+1.7\% vs.\ no embedding), attains the highest normalized reward ratio ($\approx$ 89.5\%), PPO at 128 (+0.2\%), whereas A2C performance degrades when embedding dimensions exceed zero (–0.02\% at 1536). DQN outperforms both its lower‑dimensional variants and all PPO/A2C configurations. Across all algorithms, increasing embedding size accelerates convergence and raises peak performance, underscoring the value of rich state representations in sparse‑reward RL tasks. This suggests that higher embedding sizes strike the better trade‐off between representation capacity and overfitting, with numeric state alone sufficing for simpler policy heads.

\textbf{Mouse‐Region Subdivision.}  
Extending the hierarchical agent to include multi‐step mouse‐region and sub‐region actions yields mixed performance changes (Figure~\ref{fig:hier_action}) and less reward sparsity. Compared to the no‐subregion variant, DQN’s normalized reward increases 15.5\%, PPO has 15.9\% increase, while A2C’s reward slightly decreases from 45.1\% to 41.3\% (–8.4\%). Corresponding F1 scores for the multi‐step variant are 79.2\% (DQN), 63.3\% (PPO), and 50.1\% (A2C), indicating that finer‐grained spatial control substantially benefits DQN and PPO.

\textbf{Comparison to MLLM Baselines.}  
On the OSWorld benchmark, GPT‑4o achieves a precision of 65.6\%, recall of 57.9\%, F1 of 58.0\%, and normalized reward of 35.2\%, while GPT‑4.1 improves modestly to 67.2\% precision, 61.0\% recall, 60.8\% F1, and 39.4\% reward. In contrast, our best DQN\textsubscript{CA} model attains 80.8\% precision (+15.2 over GPT‑4o), 82.0\% recall (+24.1), 81.3\% F1 (+23.3), and 89.5\% reward (+54.3), substantially outperforming both MLLM baselines. PPO\textsubscript{CA} and A2C\textsubscript{CA} also exceed GPT‑4o/GPT‑4.1 in normalized reward and most accuracy metrics. These results demonstrate that our hierarchical RL–based ComputerAgent delivers far superior task performance compared to state‑of‑the‑art multimodal LLM approaches in OSWorld with an additional 3 examples (shots).

\textbf{Summary.}
Collectively, these results demonstrate that \emph{ComputerAgent} effectively tackles the challenges outlined in section \ref{sec:intro}: \textbf{Sparse rewards \& long horizons:} Hierarchical control boosts reward by up to 50.4\,\%. \textbf{Perceptual diversity:} Triple‐modal state embedding yields up to 14.8\,\% reward improvements across OSs. \textbf{Text input \& early stopping:} Embedding task context and meta‐actions reduce wasted steps and accelerate convergence.\textbf{On‐device inference:} Instead of direct state input, embedding dimensions and a compact vision backbone ensured the entire model fits on a single consumer GPU without sacrificing performance. In addition, meta‑actions like \texttt{Wait}, \texttt{Stop}, and \texttt{Text\_input} allow the agent to adapt to tasks of varying input lengths.

\subsection{Ablation study}

\label{sec:ablation_study}

To quantify the contribution of each design choice in our ComputerAgent, we perform a three‐part ablation study on (1) the number of environment state features, (2) the embedding dimension of the action encoder, and (3) the inclusion of a streak‐bonus reward. Table~\ref{tab:ablation_combined} summarizes precision, recall, F1, and normalized reward (N.\ Rwd) for all variants.

\textbf{1. State‐Size Ablation (s6–s12)}  
As shown in Table~\ref{tab:ablation_combined} (top left), varying the state‐vector length between 6 and 12 dimensions produces only minimal performance fluctuations for DQN. F1 rises slightly from \(75.5\%\) at 6 dims to \(75.3\%\) at 8 dims, then gradually declines to \(73.0\%\) at 12 dims. Normalized reward remains essentially flat between \(83.7\%\) and \(84.0\%\) across all sizes. This flat profile reinforces that a low‐dimensional state representation already captures the core task dynamics, and adding more state variables beyond eight offers little practical benefit. For further analysis, see Appendix~\ref{app:effect_of_states}.

\vspace{0.0cm}
\begin{wraptable}{r}{0.51\textwidth} 
  \vspace{0.0cm}
  \centering
  \small
  \setlength{\tabcolsep}{1pt}      
  \renewcommand{\arraystretch}{1.1} 
  \captionsetup{skip=3pt}           

  \caption{Performance for \textbf{ComputerAgent's} ablation studies on state‑size, embedding dimension, streak bonus, and context‑aware baselines.}
  \begin{tabularx}{\linewidth}{p{15mm}XXXX}
    \toprule
    \textbf{Model} & \textbf{Prec} & \textbf{Rec} & \textbf{F1} & \textbf{N.\ Rwd} \\
    \midrule
      \multicolumn{5}{l}{\textbf{State‑Size (s6–s12)}} \\
        $DQN_{s6}$   & $74.2_{\pm0.50}$ & $77.2_{\pm0.11}$ & $75.5_{\pm0.22}$ & $83.7_{\pm1.05}$ \\
        $DQN_{s7}$   & $72.8_{\pm1.81}$ & $75.6_{\pm0.57}$ & $74.0_{\pm1.35}$ & $83.7_{\pm0.23}$ \\
        $DQN_{s8}$   & $74.3_{\pm1.59}$ & $76.7_{\pm0.03}$ & $75.3_{\pm0.94}$ & $83.7_{\pm1.01}$ \\
        $DQN_{s9}$   & $74.1_{\pm1.73}$ & $76.4_{\pm1.43}$ & $75.1_{\pm1.61}$ & $83.7_{\pm1.33}$ \\
        $DQN_{s10}$  & $72.2_{\pm3.61}$ & $74.5_{\pm3.15}$ & $73.2_{\pm3.43}$ & $83.7_{\pm1.41}$ \\
        $DQN_{s11}$  & $72.3_{\pm3.65}$ & $75.4_{\pm2.22}$ & $73.7_{\pm3.08}$ & $84.0_{\pm1.54}$ \\
        $DQN_{s12}$  & $72.0_{\pm0.41}$ & $74.3_{\pm0.20}$ & $73.0_{\pm0.31}$ & $83.5_{\pm1.05}$ \\
      \midrule
      \multicolumn{5}{l}{\textbf{Embedding Dimension}} \\
        $DQN_{256}$  & $75.7_{\pm0.74}$ & $76.7_{\pm0.39}$ & $76.1_{\pm0.58}$ & $85.1_{\pm1.11}$ \\
        $DQN_{1024}$ & $77.6_{\pm4.80}$ & $78.4_{\pm4.34}$ & $77.9_{\pm4.65}$ & $87.0_{\pm3.40}$ \\
        $DQN_{2048}$ & $76.5_{\pm1.67}$ & $78.1_{\pm1.74}$ & $77.2_{\pm1.69}$ & $86.8_{\pm2.81}$ \\
        
      \midrule
      \multicolumn{5}{l}{\textbf{W/O Streak Bonus Impact}} \\
        $DQN_{str}$  & $78.0_{\pm1.75}$ & $71.0_{\pm1.09}$ & $72.5_{\pm1.40}$ & $71.8_{\pm1.21}$ \\
        $PPO_{str}$  & $74.4_{\pm0.89}$ & $60.6_{\pm0.14}$ & $60.8_{\pm0.93}$ & $63.0_{\pm0.44}$ \\
        $A2C_{str}$  & $75.1_{\pm0.69}$ & $50.6_{\pm0.79}$ & $52.7_{\pm0.65}$ & $49.0_{\pm2.10}$ \\
      \midrule
      \multicolumn{5}{l}{\textbf{Our Model (Context‑Aware Baseline)}} \\
        $DQN_{CA}$    & $80.8_{\pm0.98}$  & $82.0_{\pm0.32}$  & $81.3_{\pm0.70}$  & $89.5_{\pm1.81}$ \\
        $PPO_{CA}$    & $67.8_{\pm0.43}$  & $67.7_{\pm2.86}$  & $65.6_{\pm1.47}$  & $65.6_{\pm5.63}$ \\
        $A2C_{CA}$    & $78.0_{\pm5.11}$  & $53.6_{\pm1.45}$  & $56.2_{\pm2.94}$  & $49.0_{\pm1.12}$ \\
    \bottomrule
  \end{tabularx}
  \vspace{-0.2cm}
  \label{tab:ablation_combined}
\end{wraptable}

\textbf{2. Embedding Dimension Ablation}  Table~\ref{tab:ablation_combined} reports DQN performance when varying the action-embedding size from 256 to 2048. All variants underperform our default 512-dimensional context-aware embedding (F1 $\approx$ 80.5\%, N.\ Rwd $\approx$ 87.5\%). Specifically, DQN$_{256}$ achieves F1 = $76.1 \pm 0.58\%$ and N.\ Rwd = $85.1 \pm 1.11\%$, DQN$_{1024}$ yields F1 = $77.9 \pm 4.65\%$ and N.\ Rwd = $87.0 \pm 3.40\%$, and DQN$_{2048}$ attains F1 = $77.2 \pm 1.69\%$ and N.\ Rwd = $86.8 \pm 2.81\%$. These modest drops of 2.6–4.4 points in F1 and up to 2.4 points in normalized reward indicate diminishing returns beyond 512 dimensions, validating our choice of a 512-dimensional embedding as both compact and expressive.

\textbf{3. Streak Bonus Impact} In Table~\ref{tab:ablation_combined}, we compare each algorithm with and without a streak‑bonus term. With streak bonuses, DQN’s F1 increases from 72.5\% to 81.3\% (+8.8\%) and normalized reward from 71.8\% to 89.5\% (+17.7\%). PPO’s F1 rises from 60.8\% to 65.6\% (+4.8\%) and reward from 63.0\% to 65.6\% (+2.6\%). A2C’s F1 improves from 52.7\% to 56.2\% (+3.5\%) with reward remaining at 49.0\%. These results indicate that streak bonuses act as an effective shaping mechanism, particularly enhancing off‑policy learning.

\section{Discovery, conclusion, limitations and future work}
\label{Limitation_and_Future_Work}

\textbf{Discovery and Conclusion} In this work, we introduce \textbf{ComputerAgent}, a hierarchical reinforcement learning framework that can be used to complete computer control tasks. Despite its lightweight footprint ($\approx$0.015 B parameters), ComputerAgent achieves an 80.8\% accuracy, surpass the 67.2\% of a 200 B‑parameter pretrained MLLM baseline—while dramatically reducing computational cost on 135 common tasks. Our qualitative analysis shows that the combination of task embeddings and carefully designed state representations substantially improves decision accuracy, and the hierarchical control structure yields further gains. We observe that DQN offers the most stable learning dynamics and the highest task‑completion accuracy, whereas PPO and A2C, though more exploratory, exhibit greater variance. Across all the agent models we have, curriculum learning has a strong effect on model performance to increase the training speed and training accuracy. We think that as the task complexity increases, the curriculum learning would have a stronger effect on helping the model to understand more sophisticated tasks. Compared to the OSWorld GPT‑4o baseline, ComputerAgent delivers competitive performance on seen tasks with far fewer resources, though its ability to generalize to novel tasks warrants future investigation. Overall, these results demonstrate that a small‐scale RL agent can rival heavyweight generative models in GUI control, paving the way for efficient on‑device deployment. In addition, a thorough discussion of these broader implications is provided in Appendix~\ref{app:Impact}.

\textbf{Limitations and Future Work} 
Despite this promise, our model was evaluated on only 135 manually designed tasks, limiting its robustness and zero‐shot generalization to unseen scenarios. For the future, we will focus on (1): Swap our dummy vision module for a lightweight, pretrained UI‑focused encoder (e.g. WideResNet-wide-50 or small vision–text transformer), fine‑tuned on GUI screenshots collected via a local emulator for on‑device deployment; (2) applying meta‐ and self‐supervised learning to improve transfer to novel tasks, and (3) scaling our evaluation to a broader, more diverse task suite. (4) Implementing continuous control for more accurate mouse control. Although considerations of social impact, ethics, safeguards, and licensing are essential for the responsible deployment of AI systems, they do not influence our model’s performance metrics or experimental outcomes.







\newpage
\bibliographystyle{unsrt}
\bibliography{ref} 

\begin{thebibliography}{10}

\bibitem{schulman2017proximal}
John Schulman, Filip Wolski, Prafulla Dhariwal, Alec Radford, and Oleg Klimov.
\newblock Proximal policy optimization algorithms.
\newblock {\em arXiv preprint arXiv:1707.06347}, 2017.

\bibitem{lin2014microsoft}
Tsung-Yi Lin, Michael Maire, Serge Belongie, James Hays, Pietro Perona, Deva Ramanan, Piotr Doll{\'a}r, and C~Lawrence Zitnick.
\newblock Microsoft coco: Common objects in context.
\newblock In {\em Computer vision--ECCV 2014: 13th European conference, zurich, Switzerland, September 6-12, 2014, proceedings, part v 13}, pages 740--755. Springer, 2014.

\bibitem{xie2024osworldbenchmarkingmultimodalagents}
Tianbao Xie, Danyang Zhang, Jixuan Chen, Xiaochuan Li, Siheng Zhao, Ruisheng Cao, Toh~Jing Hua, Zhoujun Cheng, Dongchan Shin, Fangyu Lei, Yitao Liu, Yiheng Xu, Shuyan Zhou, Silvio Savarese, Caiming Xiong, Victor Zhong, and Tao Yu.
\newblock Osworld: Benchmarking multimodal agents for open-ended tasks in real computer environments, 2024.

\bibitem{lee2025efficient}
Jewon Lee, Ki-Ung Song, Seungmin Yang, Donguk Lim, Jaeyeon Kim, Wooksu Shin, Bo-Kyeong Kim, Yong~Jae Lee, and Tae‑Ho Kim.
\newblock {Efficient LLaMA-3.2-Vision by Trimming Cross-attended Visual Features}, 2025.
\newblock Accepted at CVPR 2025 Workshop on ELVM.

\bibitem{humphreys2022learning}
Peter~C. Humphreys, David Raposo, Toby Pohlen, Gregory Thornton, Rachita Chhaparia, Alistair Muldal, Josh Abramson, Petko Georgiev, Alex Goldin, Adam Santoro, and Timothy Lillicrap.
\newblock A data-driven approach for learning to control computers.
\newblock In {\em Proceedings of the 39th International Conference on Machine Learning (ICML)}, volume 162, pages 9153--9166. PMLR, 2022.

\bibitem{deng2023mind2web}
Xiang Deng, Yu~Gu, Boyuan Zheng, Shijie Chen, Samuel Stevens, Boshi Wang, Huan Sun, and Yu~Su.
\newblock {Mind2Web}: Towards a generalist agent for the web.
\newblock In {\em Advances in Neural Information Processing Systems 36 (NeurIPS 2023), Datasets and Benchmarks Track}, 2023.

\bibitem{nakano2021webgpt}
Reiichiro Nakano, Jacob Hilton, Suchir Balaji, Jeff Wu, Long Ouyang, Christina Kim, Christopher Hesse, Shantanu Jain, Vik~Goel Kosaraju, William Saunders, et~al.
\newblock {WebGPT}: Browser-assisted question-answering with human feedback.
\newblock 2021.

\bibitem{gao2024assistgui}
Difei Gao, Lei Ji, Zechen Bai, Mingyu Ouyang, Peiran Li, Dongxing Mao, Qinchen Wu, Weichen Zhang, Peiyi Wang, Xiangwu Guo, Hengxu Wang, Luowei Zhou, and Mike~Zheng Shou.
\newblock {AssistGUI}: Task-oriented desktop graphical user interface automation.
\newblock In {\em Proceedings of the IEEE/CVF Conference on Computer Vision and Pattern Recognition (CVPR)}, 2024.

\bibitem{zhang2024ufo}
Chaoyun Zhang, Liqun Li, Shilin He, Xu~Zhang, Bo~Qiao, Si~Qin, Minghua Ma, Yu~Kang, Qingwei Lin, Saravan Rajmohan, Dongmei Zhang, and Qi~Zhang.
\newblock {UFO}: A {UI}-focused agent for windows os interaction.
\newblock {\em arXiv preprint arXiv:2402.07939}, 2024.

\bibitem{tan2024cradle}
Weihao Tan, Wentao Zhang, Xinrun Xu, Haochong Xia, Ziluo Ding, Boyu Li, Bohan Zhou, Junpeng Yue, Jiechuan Jiang, Yewen Li, Ruyi An, Molei Qin, Chuqiao Zong, Longtao Zheng, Yujie Wu, Xiaoqiang Chai, Yifei Bi, Tianbao Xie, Pengjie Gu, Xiyun Li, Ceyao Zhang, Long Tian, Chaojie Wang, Xinrun Wang, Börje~F. Karlsson, Bo~An, Shuicheng Yan, and Zongqing Lu.
\newblock {Cradle}: Empowering foundation agents towards general computer control.
\newblock {\em arXiv preprint arXiv:2403.03186}, 2024.

\bibitem{wu2024oscopilot}
Zhiyong Wu, Chengcheng Han, Zichen Ding, Zhenmin Weng, Zhoumianze Liu, Shunyu Yao, Tao Yu, and Lingpeng Kong.
\newblock {OS-Copilot}: Towards generalist computer agents with self-improvement.
\newblock {\em arXiv preprint arXiv:2402.07456}, 2024.

\bibitem{lu2024omniparserpurevisionbased}
Yadong Lu, Jianwei Yang, Yelong Shen, and Ahmed Awadallah.
\newblock Omniparser for pure vision based gui agent, 2024.

\bibitem{Sutton1999Options}
Richard~S. Sutton, Doina Precup, and Satinder Singh.
\newblock Between {MDPs} and {SMDPs}: A framework for temporal abstraction.
\newblock {\em Artificial Intelligence}, 112(1--2):181--211, 1999.

\bibitem{Bacon2017OptionCritic}
Pierre{-}Luc Bacon, Jean Harb, and Doina Precup.
\newblock The option{-}critic architecture.
\newblock In {\em AAAI}, 2017.

\bibitem{Kulkarni2016HDQN}
Tejas~D. Kulkarni, Karthik Narasimhan, Ardavan Saeedi, and Joshua~B. Tenenbaum.
\newblock Hierarchical deep reinforcement learning: Integrating temporal abstraction and intrinsic motivation.
\newblock In {\em NeurIPS}, 2016.

\bibitem{Vezhnevets2017Feudal}
Alexander Sasha~Vezhnevets et~al.
\newblock Feudal networks for hierarchical reinforcement learning.
\newblock In {\em ICML}, 2017.

\bibitem{mnih2013playing}
Volodymyr Mnih, Koray Kavukcuoglu, David Silver, Alex Graves, Ioannis Antonoglou, Daan Wierstra, and Martin Riedmiller.
\newblock Playing atari with deep reinforcement learning.
\newblock {\em arXiv preprint arXiv:1312.5602}, 2013.

\bibitem{Nachum2018HIRO}
Ofir Nachum, Shixiang Gu, Honglak Lee, and Sergey Levine.
\newblock Data{-}efficient hierarchical reinforcement learning.
\newblock In {\em NeurIPS}, 2018.

\bibitem{Florensa2017SNN}
Carlos Florensa, Yan Duan, and Pieter Abbeel.
\newblock Stochastic neural networks for hierarchical reinforcement learning.
\newblock In {\em ICLR}, 2017.

\bibitem{Gurtler2021HiTS}
Julius Gürtler and Andreas Martius.
\newblock Hits: Hierarchical reinforcement learning with timed subgoals.
\newblock {\em arXiv preprint arXiv:2103.01265}, 2021.

\bibitem{Pateria2021Survey}
Saurabh Pateria, Akshat Kumar, Jilles~S. Dibangoye, and Balaraman Ravindran.
\newblock Hierarchical reinforcement learning: A comprehensive survey.
\newblock {\em ACM Computing Surveys}, 2021.

\bibitem{Narvekar2020CurriculumSurvey}
Sanmit~Narvekar et~al.
\newblock Curriculum learning for reinforcement learning domains: A framework and survey.
\newblock {\em Journal of Machine Learning Research}, 21:1--50, 2020.

\bibitem{Jain2022UIHRL}
Ankesh Jain, Yanbin Liu, and Pieter Abbeel.
\newblock Learning to interact with a complex interface using hierarchical reinforcement learning.
\newblock {\em arXiv preprint arXiv:2204.10374}, 2022.

\bibitem{bacon2017option}
Pierre-Luc Bacon, Jean Harb, and Doina Precup.
\newblock The option-critic architecture.
\newblock In {\em Proceedings of the 31st AAAI Conference on Artificial Intelligence (AAAI)}, pages 1726--1734, 2017.

\bibitem{Harb2018TerminationCritic}
Jean Harb, Pierre{-}Luc Bacon, Georgios Konidaris, and Doina Precup.
\newblock When waiting is not an option: Learning options with a deliberation cost.
\newblock {\em AAAI}, 2018.

\bibitem{Jiang2019LearningToStop}
Shengyi Jiang, Alex Kulesza, and Satinder Singh.
\newblock Learning to stop: A simple sufficient condition for convergence of stochastic adaptive control algorithms.
\newblock {\em arXiv preprint arXiv:1904.08133}, 2019.

\bibitem{Florensa2017ReverseCurriculum}
Carlos Florensa, David Held, Markus Wulfmeier, Michael Zhang, and Pieter Abbeel.
\newblock Reverse curriculum generation for reinforcement learning.
\newblock In {\em CoRL}, 2017.

\bibitem{Matiisen2020TeacherStudent}
Tambet Matiisen, Oliver~P. Sela, Raul Vicente, and Juergen Schmidhuber.
\newblock Teacher–student curriculum learning.
\newblock {\em IEEE Transactions on Neural Networks and Learning Systems}, 2020.

\bibitem{he2016deep}
Kaiming He, Xiangyu Zhang, Shaoqing Ren, and Jian Sun.
\newblock Deep residual learning for image recognition.
\newblock In {\em Proceedings of the IEEE conference on computer vision and pattern recognition}, pages 770--778, 2016.

\bibitem{OpenAI_Introducing_Operator}
OpenAI.
\newblock Introducing operator.
\newblock \url{https://openai.com/index/introducing-operator/}.
\newblock Accessed: 2025-04-08.

\bibitem{jolliffe2002principal}
Ian~T. Jolliffe.
\newblock {\em Principal Component Analysis}.
\newblock Springer Series in Statistics, 2 edition, 2002.

\bibitem{mnih2016asynchronous}
Volodymyr Mnih, Adria~Puigdomenech Badia, Mehdi Mirza, Alex Graves, Timothy Lillicrap, Tim Harley, David Silver, and Koray Kavukcuoglu.
\newblock Asynchronous methods for deep reinforcement learning.
\newblock In {\em International conference on machine learning}, pages 1928--1937. PmLR, 2016.

\end{thebibliography}

\clearpage
\appendix
\section*{Appendix}

\section{Value-Based vs. Policy-Based Learning Objectives}
\label{appendix:vb_pb_learning_obj}

\begin{table}[H]
\centering
\caption{Value‐ vs. Policy‐based objectives used in our hierarchy.  For brevity, minibatch index \(i\) is omitted.  \(\hat A\) is the advantage, \(R\) the \(n\)-step return.}
\label{tab:vb_pb_losses}
\begin{tabular}{p{2.0cm} p{7.2cm} p{4.5cm}}
\toprule
\textbf{Algorithm} & \textbf{Loss (per transition)} & \textbf{Notes} \\
\midrule
\textbf{DQN} &
\( L_{\text{DQN}}
  =\bigl[r + \gamma \max_{a'} Q_{\!t}(s',a') - Q(s,a)\bigr]^2 \) &
Value-based; discrete actions, replay buffer, target network. \\
\addlinespace
\textbf{PPO} &
\( L_{\text{PPO}}
  = \min\!\Bigl( r\,\hat A,\;
        \mathrm{clip}\!\bigl(r,\;1\!\pm\!\epsilon\bigr)\,\hat A
      \Bigr),\;
   r = \tfrac{\pi_\theta(a\mid s)}{\pi_{\theta_{\text{old}}}(a\mid s)} \) &
Clipped policy gradient; smooth updates for continuous cursor control. \\
\addlinespace
\textbf{A2C} &
\( L_{\text{A2C}}
   = -\log\pi_\theta(a\mid s)\,\hat A
     + \tfrac{1}{2}\!\bigl[\,V_\theta(s) - R\,\bigr]^2 \) &
Synchronous actor–critic; low-latency updates for changing sub-goals. \\
\bottomrule
\end{tabular}
\[
\hat A = r + \gamma V_\theta(s') - V_\theta(s), 
\qquad
R = \sum_{k=0}^{n-1} \gamma^{k} r_{t+k} + \gamma^{n} V_\theta(s_{t+n})
\]
\end{table}

\section{Agent actions and descriptions}

In our hierarchical RL framework, the agent operates by invoking a fixed set of low‑level action primitives, each corresponding to a single “manager” in the controller. Table~\ref{tab:functions_grouped} lists representative examples from each of the four manager categories:

\begin{itemize}[leftmargin=*]
  \item \textbf{Single‑Key Actions}: Direct key‐presses of individual characters. These are used for text entry or simple keystrokes (e.g., \texttt{single\_key(`a`)}).
  \item \textbf{Hotkey Actions}: Combination key‐presses (e.g., copy/paste or window‑switching). Hotkeys allow the agent to invoke common OS shortcuts in one atomic step (e.g., \texttt{hotkey(`ctrl+c`)}).
  \item \textbf{Meta‑Key Actions}: High‑level control signals such as waiting or branching into an LLM for text input. These primitives enable the agent to pause for external input (e.g., \texttt{meta\_key(`wait`)}) or to terminate its action sequence (e.g., \texttt{meta\_key(`stop`)}).
  \item \textbf{Mouse Control}: Fine‑grained pointer operations parameterized by region and subregion indices along with the interaction type (press, release, scroll). For example, \texttt{mouse\_control(r=1, subr=2, `left\_press`)} moves the cursor into region 1/subregion 2 and depresses the left button.
\end{itemize}

\label{Agent_Function_Examples}
\begin{table}[H]
\centering
\begin{tabular}{@{} p{5cm} p{8cm} @{}}
\toprule
\textbf{Function} & \textbf{Description} \\
\midrule
\multicolumn{2}{l}{\textbf{Single‐Key Actions}} \\
single\_key(`a`)  & Presses the “a” key. \\
single\_key(`k`)  & Presses the “k” key. \\
\midrule
\multicolumn{2}{l}{\textbf{Hotkey Actions}} \\
hotkey(`ctrl+c`)  & Performs the Ctrl+C combination (copy). \\
hotkey(`alt+tab`) & Performs the Alt+Tab combination (window switch). \\
\midrule
\multicolumn{2}{l}{\textbf{Meta‐Key Actions}} \\
meta\_key(`wait`)            & Agent pauses execution (wait). \\
meta\_key(`llm\_text\_input`) & Agent invokes LLM for text input. \\
\midrule
\multicolumn{2}{l}{\textbf{Mouse Control}} \\
mouse\_control(r=1, subr=2, `left\_press`)    & Left‐button press in region 1, subregion 2. \\
mouse\_control(r=3, subr=6, `scroll\_release`) & Scroll‐release in region 3, subregion 6. \\
\bottomrule
\end{tabular}
\caption{Example function calls for each action‐manager category with brief descriptions.}
\label{tab:functions_grouped}
\end{table}

\section{Hierarchical Action Space}
We defined number of action spaces for Manager Agent and Action Agents for each tasks below.
\label{appendix:hier_action}
\begin{table}[ht]
\centering
\begin{tabular}{@{} l l c @{}}
\toprule
\textbf{Category}      & \textbf{Description}                                                      & \textbf{\#Actions} \\
\midrule
Single Keys            & Individual keyboard buttons (e.g., \texttt{Esc}, \texttt{Enter}, \texttt{Space}, $\ldots$)       & 92             \\
Composite Keys         & Hotkeys and predefined commands (e.g., \texttt{Ctrl+X}, \texttt{Shift+X}, \texttt{Cmd+X}, $\ldots$) & 78             \\
Meta Actions           & High‐level task signals (\texttt{Start}, \texttt{Stop}, \texttt{Wait}, \texttt{Text\_Field\_Input}) & 4              \\
\midrule
\multirow{3}{*}{Mouse Control}
                       & Coarse region selection sub‐policy                                       & 9              \\
                       & Fine‐grained subregion selection sub‐policy                              & 9              \\
                       & Interaction types (e.g., \texttt{left\_press}, \texttt{left\_release}, $\ldots$) & 8              \\
\bottomrule
\end{tabular}
\caption{Breakdown of the agent’s action space.}
\label{tab:action-space}
\end{table}

\section{Penalty instantiation}
\label{app:Penalty}
\paragraph{Penalty instantiation.}
With default hyper‑parameters, the additive penalty~\eqref{eq:penalty} becomes
\(
P = P_{\mathrm{repeat}}
   + P_{\mathrm{pointer}}
   + P_{\mathrm{step}}
   + P_{\mathrm{early}}
   + P_{\mathrm{exp}} .
\)
Each component is defined as follows.  

\begin{enumerate}[leftmargin=1.2em,itemsep=4pt]
\item \emph{Repeat penalty.}  
      Let \(n\) be the length of the current repetition streak and
      \(n_0\!=\)\texttt{repeat\_threshold}${}=2$:
      \[
        P_{\mathrm{repeat}}
        \;=\;
        3.0\,\bigl(2^{\,n-n_0}-1\bigr)\,
        \mathbb{1}[\,n>n_0\,].
      \]

\item \emph{Pointer‑stagnation penalty.}  
      If the task pointer has not advanced for
      \(c>\!3\) consecutive steps,
      \[
        P_{\mathrm{pointer}}
        \;=\;
        2.0\,\mathbb{1}[\,c>3\,],
      \qquad
        c:=\texttt{unchanged\_count}.
      \]

\item \emph{Step penalty.}  
      A constant cost promotes concise action sequences:
      \[
        P_{\mathrm{step}} \;=\; 0.05 .
      \]

\item \emph{Early‑termination penalty.}  
      Denote the agent’s progress pointer by \(p\) and the length of the
      ground‑truth sequence by \(L\):
      \[
        P_{\mathrm{early}}
        \;=\;
        2.0\,\mathbb{1}[\,p<L\;\wedge\;\va_t^{\mathrm{type}}=\textsc{Stop}\,].
      \]

\item \emph{Exploration penalty.}  
      Let \(t\) be the current timestep and
      \(T^\star\) the task’s nominal horizon:
      \[
        P_{\mathrm{exp}}
        \;=\;
        0.2\;\bigl(1.1\bigr)^{\max(0,\,t-T^\star)} .
      \]
\end{enumerate}

\section{Experiment}

\subsection{Data Collection and Cleaning}
\label{app:Data}

We manually designed a suite of 135 GUI tasks by surveying common real‑world operations—spanning OS control, web browsing, file management, and system settings—each defined as a single‑goal problem with one optimal action sequence. Tasks were classified as simple (action length < 8) or hard (action length $\geq$ 8) to ensure coverage across difficulty levels.

To guarantee annotation quality, each author independently reviewed every task. Disagreements were flagged and resolved through group discussion. Once consensus was reached, we ran an automated validation script that “replayed” each task against our defined action space, verifying success and flagging any sequences that failed to execute. Tasks that did not pass validation were corrected.

Finally, we performed a cleaning pass to remove duplicates, normalize action names, and check for consistency in formatting. This rigorous process ensures that our 135‑task benchmark is both representative of real‑world GUI use cases and fully executable within the ComputerAgent framework.

\subsection{Experimental Setup Details}
\label{app:Exp_setup}

\begin{table}[H]
\centering
\caption{Experimental setup hyperparameters}
\label{tab:exp_setup_details}
\begin{tabular}{@{}ll@{}}
\toprule
\multicolumn{2}{l}{\textbf{Main}} \\
\midrule
Max steps                        & 100      \\
Target update interval           & 50       \\
Num episodes                     & 100000   \\
\midrule
\multicolumn{2}{l}{\textbf{Environment}} \\
\midrule
$\alpha$                         & 1.2      \\
manager\_correct\_reward         & 2        \\
subpolicy\_correct\_reward       & 6        \\
manager\_streak\_bonus           & 2        \\
subpolicy\_streak\_bonus         & 2        \\
base\_step\_penalty              & -0.5     \\
exp\_penalty\_base               & -0.2     \\
exp\_penalty\_factor             & 1.1      \\
short\_ending\_penalty           & -2.0     \\
pointer\_unchanged\_threshold    & 3        \\
pointer\_unchanged\_penalty      & -4.0     \\
repeat\_threshold                & 2        \\
repeat\_exp\_base                & -3.0     \\
repeat\_exp\_factor              & 2.0      \\
mouse\_region\_reward            & 3        \\
mouse\_interaction\_reward       & 3        \\
negative\_stop\_threshold        & -200     \\
\midrule
\multicolumn{2}{l}{\textbf{Epsilon (DQN only)}} \\
\midrule
epsilon\_start                   & 1.0      \\
epsilon\_end                     & 0.007    \\
epsilon\_decay                   & 8000     \\
\midrule
\multicolumn{2}{l}{\textbf{DQN Network}} \\
\midrule
FC dimensions                    & [512, 512, 512] \\
\bottomrule
\end{tabular}
\end{table}

As Section~\ref{Experiment_settings} mentions, we employ a unified set of hyperparameters to control the training regimen, environment dynamics, exploration schedule, and network architecture across all algorithms. Table~\ref{tab:exp_setup_details} provides a comprehensive overview of these settings, including main loop parameters, reward shaping coefficients, exploration schedule for DQN, and the fully connected network dimensions.

\subsection{Detailed Metric Definitions}
\label{app:metrics}

We evaluate our agent’s predicted action sequence against the ground‐truth script of each task. Let: $\mathrm{TP}$ be the number of atomic GUI actions the agent executes that match the corresponding ground‐truth actions, $\mathrm{FP}$ be the number of extra (incorrect) actions the agent issues, and $\mathrm{FN}$ be the number of ground‐truth actions the agent fails to issue. We then compute:
\[
  \mathrm{Precision} = \frac{\mathrm{TP}}{\mathrm{TP} + \mathrm{FP}},
  \quad
  \mathrm{Recall}    = \frac{\mathrm{TP}}{\mathrm{TP} + \mathrm{FN}},
  \quad
  F_1 = 2 \cdot \frac{\mathrm{Precision} \times \mathrm{Recall}}{\mathrm{Precision} + \mathrm{Recall}}.
\]

\subsection{Computational resource Settings}
\label{app:infrastructure}
\textbf{Intrastrcture-wise}, As requested for replication, our experiment is running on the Lenovo P8 with AMD Ryzen Threadripper PRO 7995WX 2.5GHz 96-core with 128GB memory and NVIDIA RTX A6000 running on Ubuntu 22.04. With 43 set of experiments, each experiment will need on average 4 hours to run all 100k episodes. However, this does not means the model we trained will need that much computational power, \textbf{it is still eligible for running on local small device.}

\subsection{Main Result: Step‐Count Robustness}
\label{app:main_result}

\begin{figure}[H]
    \centering
    \includegraphics[width=0.7\linewidth]{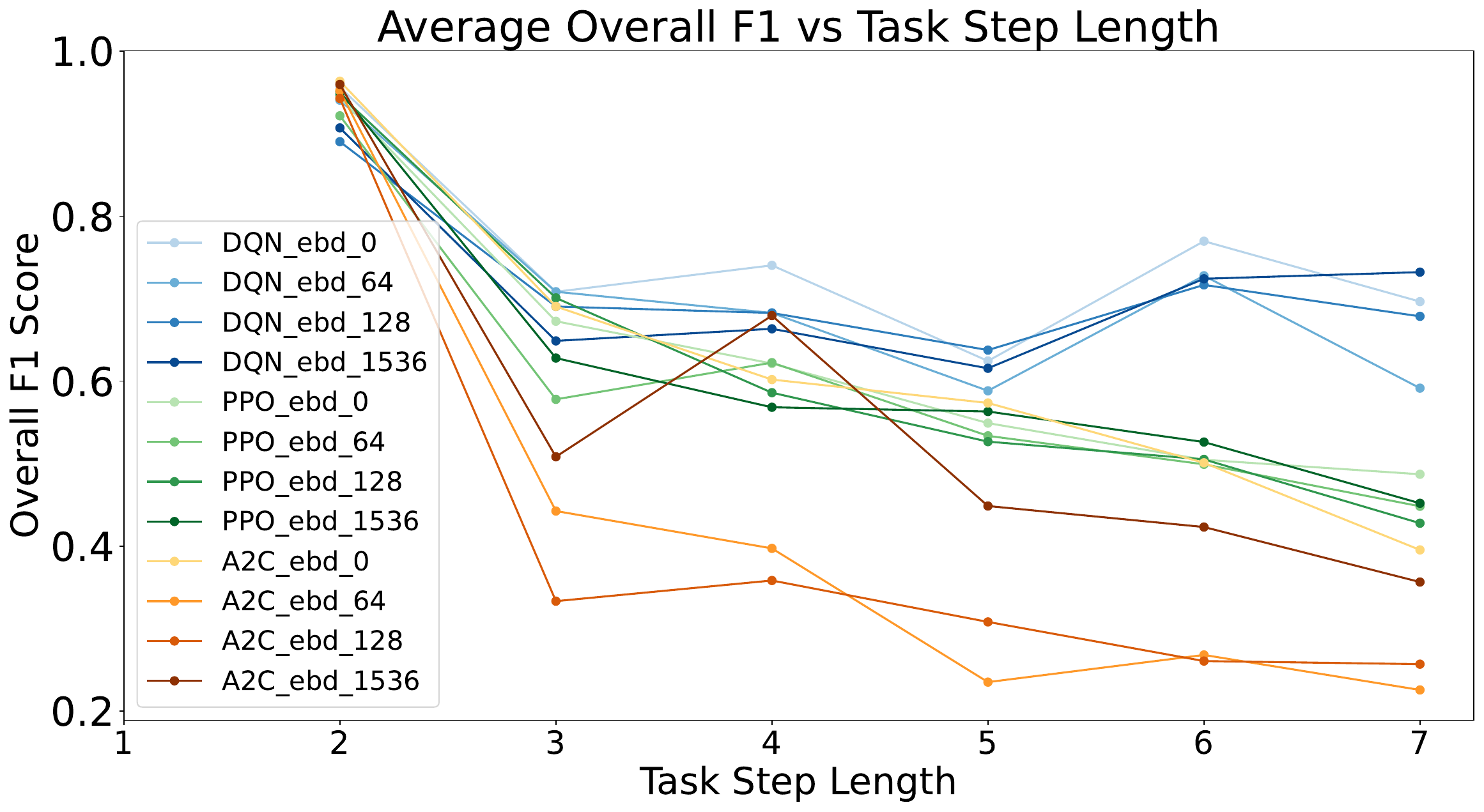}
    \caption{Overall F1 score as a function of task step length for DQN, PPO, and A2C with embedding dimensions of 0, 64, 128, and 1536, showing that richer embeddings sustain high accuracy on longer‐horizon tasks, whereas smaller or absent embeddings suffer steep performance drops as task length grows.}
    \label{fig:f1_step}
\end{figure}

\begin{figure*}[htbp]
  \centering
  \begin{subfigure}[b]{0.33\linewidth}
    \includegraphics[width=\linewidth]{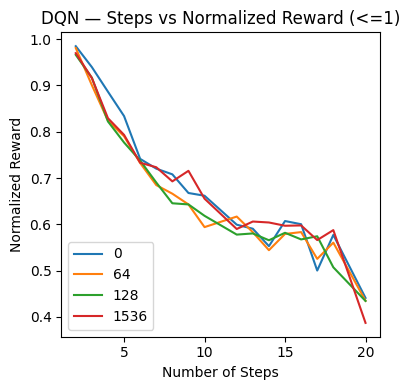}
    \caption{DQN: normalized reward vs.\ steps}
    \label{fig:DQN_rwd_stp}
  \end{subfigure}\hspace{0pt}%
  \begin{subfigure}[b]{0.33\linewidth}
    \includegraphics[width=\linewidth]{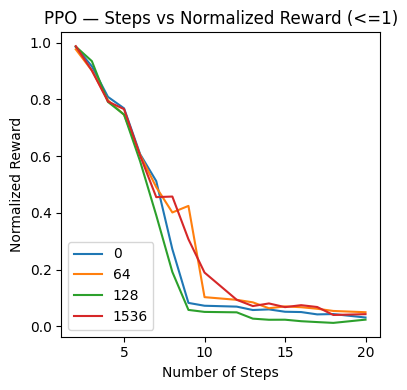}
    \caption{PPO: normalized reward vs.\ steps}
    \label{fig:PPO_rwd_stp}
  \end{subfigure}\hspace{0pt}%
  \begin{subfigure}[b]{0.33\linewidth}
    \includegraphics[width=\linewidth]{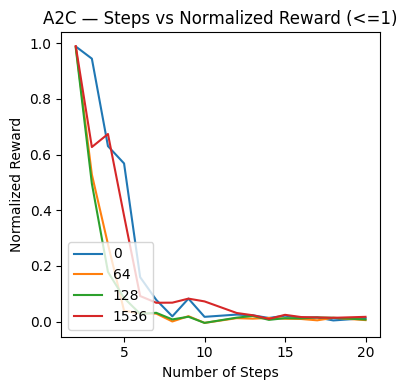}
    \caption{A2C: normalized reward vs.\ steps}
    \label{fig:A2C_rwd_stp}
  \end{subfigure}
  \caption{\textbf{Normalized reward as a function of task length.} Subplots (a), (b), and (c) show DQN, PPO, and A2C, respectively, with curves for embedding dimensions of 0, 64, 128, and 1536. Rewards are averaged across tasks grouped by the number of steps required, demonstrating how longer sequences challenge each algorithm and how richer embeddings can partially offset this effect.}
  \label{fig:rwd_stp}
\end{figure*}

Beyond our main study, we evaluate how task length affects each agent from both Normalized Rewards and F1 Scores (Table~\ref{tab:stepcat_mixed} and Figure~\ref {fig:f1_step},~\ref{fig:rwd_stp}). On “simple” tasks (step‑count<8), all variants achieve high performance (NormRew \(\ge 0.578\), O‑F1 \(\ge 0.641\)). However, on “hard” tasks (step‑count \(\ge 8\)), on‑policy methods collapse: PPO’s NormRew falls to 0.071–0.107 (O‑F1 0.274–0.412) and A2C’s to 0.010–0.027 (O‑F1 0.182–0.246). In contrast, DQN remains robust, with NormRew 0.568–0.596 and O‑F1. 

\begin{table}[H]
  \centering
  \setlength{\tabcolsep}{2pt}
  \renewcommand{\arraystretch}{1.3}
  \begin{tabular}{lrrrrrrrr}
    \toprule
    \textbf{Model}    & \multicolumn{4}{c}{\textbf{Simple (steps<8)}} & \multicolumn{4}{c}{\textbf{Hard (steps\(\ge\)8)}} \\
                      & \textbf{NormRew} & \textbf{O-F1} & \textbf{O-Prec} & \textbf{O-Rec} & \textbf{NormRew} & \textbf{O-F1} & \textbf{O-Prec} & \textbf{O-Rec} \\
    \midrule
    $DQN_{0}$     & $91.0_{\pm0.71}$ & $84.2_{\pm0.73}$ & $84.0_{\pm0.71}$ & $84.4_{\pm0.75}$ & $58.9_{\pm0.05}$  & $70.3_{\pm2.90}$  & $69.8_{\pm2.63}$  & $70.9_{\pm3.21}$ \\
    $DQN_{64}$    & $91.7_{\pm0.61}$ & $83.8_{\pm1.48}$ & $83.3_{\pm2.01}$ & $84.7_{\pm0.51}$ & $57.6_{\pm0.95}$  & $68.4_{\pm0.13}$  & $67.7_{\pm0.01}$  & $69.2_{\pm0.27}$ \\
    $DQN_{128}$   & $91.7_{\pm0.71}$ & $83.8_{\pm1.76}$ & $83.3_{\pm1.87}$ & $84.6_{\pm1.54}$ & $55.8_{\pm4.06}$  & $68.1_{\pm1.64}$  & $67.1_{\pm2.62}$  & $69.4_{\pm0.30}$ \\
    $DQN_{1536}$  & $92.1_{\pm0.97}$ & $85.8_{\pm0.08}$ & $85.4_{\pm0.29}$ & $86.3_{\pm0.27}$ & $60.8_{\pm2.60}$  & $72.2_{\pm1.39}$  & $71.6_{\pm1.80}$  & $72.9_{\pm0.89}$ \\
    \midrule
    $PPO_{0}$     & $85.3_{\pm0.61}$ & $76.5_{\pm0.31}$ & $74.3_{\pm0.46}$ & $80.4_{\pm0.23}$ & $ 8.62_{\pm2.12}$ & $37.5_{\pm2.90}$  & $64.9_{\pm4.24}$  & $29.2_{\pm4.37}$ \\
    $PPO_{64}$    & $85.2_{\pm0.65}$ & $75.4_{\pm1.48}$ & $72.3_{\pm1.96}$ & $80.4_{\pm0.41}$ & $12.1_{\pm2.10}$  & $41.5_{\pm2.21}$  & $56.9_{\pm3.66}$  & $37.4_{\pm5.93}$ \\
    $PPO_{128}$   & $84.9_{\pm0.55}$ & $75.4_{\pm0.33}$ & $72.7_{\pm0.64}$ & $80.4_{\pm0.03}$ & $ 7.97_{\pm5.54}$ & $34.6_{\pm10.30}$ & $68.8_{\pm17.20}$ & $27.3_{\pm13.20}$ \\
    $PPO_{1536}$  & $85.7_{\pm0.67}$ & $76.7_{\pm0.94}$ & $74.4_{\pm1.32}$ & $80.3_{\pm0.10}$ & $12.3_{\pm2.12}$  & $42.9_{\pm2.32}$  & $54.5_{\pm4.33}$  & $41.7_{\pm8.36}$ \\
    \midrule
    $A2C_{0}$     & $70.0_{\pm3.69}$ & $72.5_{\pm6.18}$ & $76.3_{\pm7.55}$ & $72.9_{\pm3.90}$ & $ 1.88_{\pm0.28}$ & $22.1_{\pm1.32}$  & $13.0_{\pm0.99}$  & $13.0_{\pm0.99}$ \\
    $A2C_{64}$    & $64.7_{\pm7.63}$ & $67.8_{\pm3.92}$ & $74.6_{\pm1.79}$ & $67.9_{\pm5.32}$ & $ 1.29_{\pm0.43}$ & $19.2_{\pm1.32}$  & $76.0_{\pm6.85}$  & $11.0_{\pm0.71}$ \\
    $A2C_{128}$   & $64.9_{\pm10.10}$& $67.5_{\pm4.76}$ & $75.5_{\pm2.64}$ & $67.1_{\pm6.73}$ & $ 2.11_{\pm1.10}$ & $22.0_{\pm2.94}$  & $74.8_{\pm6.07}$  & $13.1_{\pm2.26}$ \\
    $A2C_{1536}$  & $71.3_{\pm2.82}$ & $72.6_{\pm2.75}$ & $76.1_{\pm5.87}$ & $73.2_{\pm0.97}$ & $ 2.15_{\pm0.82}$ & $22.4_{\pm3.16}$  & $83.0_{\pm3.08}$  & $13.2_{\pm2.35}$ \\
    \bottomrule

  \end{tabular}
  \caption{Performance metrics of DQN, PPO, and A2C variants across simple (step‑count $<8$) and hard (step‑count \(\ge 8\)) task categories.}
  \label{tab:stepcat_mixed}
\end{table}

\section{Abliation Study}

\subsection{Effect of State‐Space Dimensionality on Performance}
\label{app:effect_of_states}
\begin{figure}[H]
    \centering
    \includegraphics[width=0.6\linewidth]{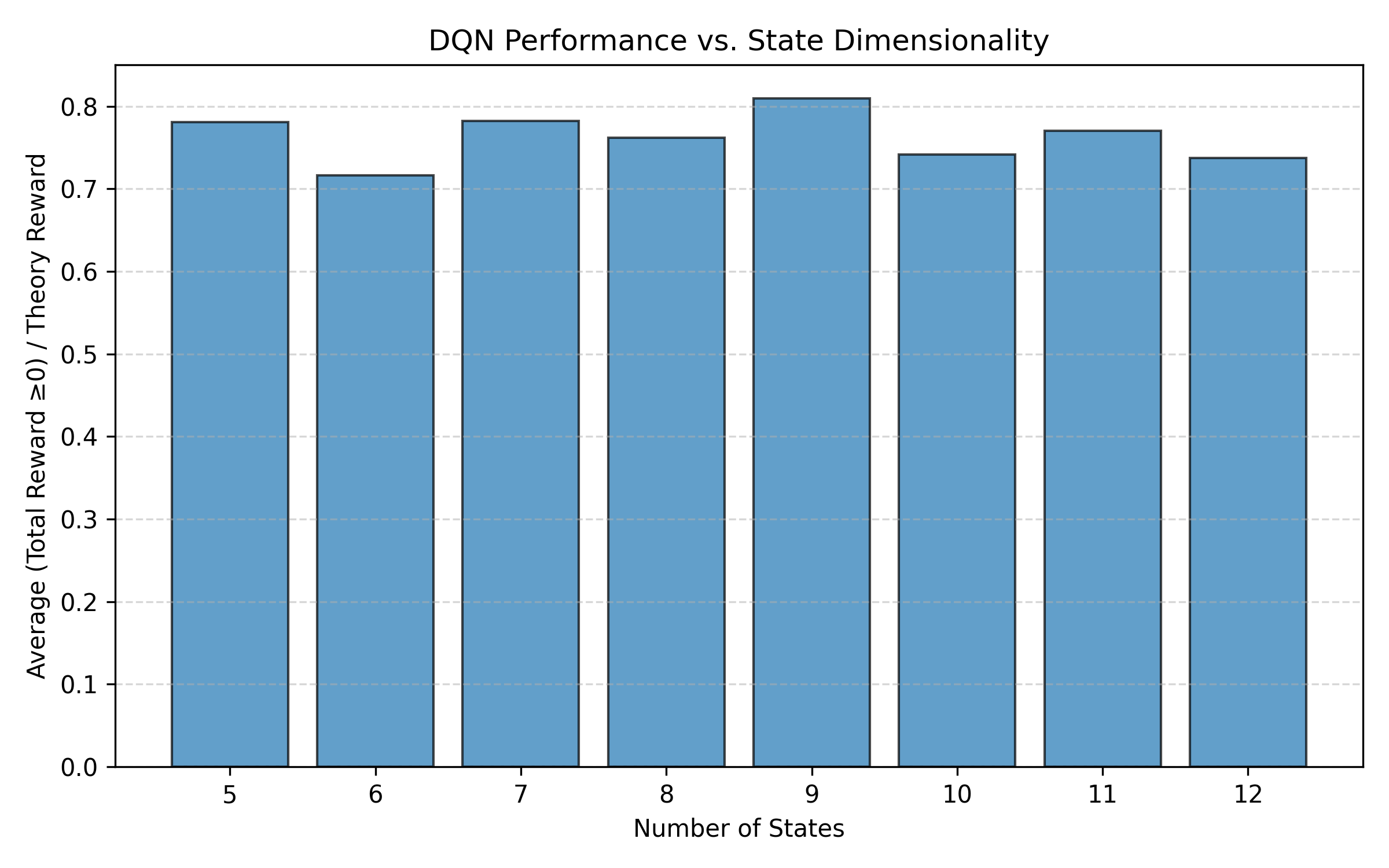}
    \caption{\textbf{Impact of state‐space dimensionality on DQN reward.} Bars show the mean normalized reward across all tasks for each state‐vector length (5–12). The relatively flat profile indicates only marginal benefit from increasing the number of state features.}
    \label{fig:state_dim}
\end{figure}

Figure~\ref{fig:state_dim} reports the average normalized reward achieved by the DQN agent as we vary the number of state features from 5 to 12. Although minor fluctuations occur (e.g., a slight peak at 9 dimensions), overall performance remains largely unchanged across this range. This stability indicates that adding extra state variables beyond our default five does not yield substantial gains, suggesting that the core task dynamics are already well captured by a low‐dimensional representation.

\section{Social Impacts}
\label{app:Impact}
The Computer Agent is confined to the 135 manually curated tasks defined in Section~\ref{sec:intro}, with no capability to access files, networks, or processes beyond those tasks.  Under normal operation, there is no direct path to large‑scale harm or unintended system manipulation.

At the same time, by uniting hierarchical reinforcement learning with natural‑language–driven control and LLM reasoning, our framework can significantly broaden access to GUI automation.  Users with limited mobility or those requiring hands‑free operation can issue high‑level instructions in plain language, while the agent executes low‑level GUI interactions on their behalf, potentially increasing productivity across a wide range of applications.

However, any general‑purpose automation tool may be misused if deployed without safeguards.  To mitigate risk, we recommend: (1) sandboxed execution environments that restrict the agent to pre‑approved applications; (2) human‑in‑the‑loop approval of unfamiliar or high‑impact action sequences; and (3) comprehensive logging and anomaly detection to flag unexpected behaviors.  

All software components and task definitions in this project are original creations of the authors, and we release them under an open‑source license.  No third‑party code, data, or models are incorporated, and the 135‑task evaluation suite is provided alongside our code to ensure full transparency and reproducibility.

\end{document}